\documentclass[journal]{IEEEtran}

\ifCLASSINFOpdf
\else
\fi

\usepackage{graphicx}
\usepackage{amssymb}
\usepackage{epstopdf}
\usepackage{subfigure}
\usepackage{algorithm}
\usepackage{algorithmic}
\usepackage{amsmath}
\usepackage{enumerate}
\usepackage{color}
\usepackage{textcomp}

\newtheorem{proposition}{Proposition}

\newtheorem{definition}{Definition}
\newtheorem{problem}{Problem}

\newcommand{\revis}[1]{#1}
\newcommand{\revistwo}[1]{#1}
\newcommand{\revisthree}[1]{#1}

\begin{document}

\title{Learning Smooth Pattern Transformation Manifolds}

\author{Elif Vural and Pascal Frossard
\thanks{E. Vural and P. Frossard are with Ecole Polytechnique F\'{e}d\'{e}rale de Lausanne (EPFL), Signal Processing Laboratory - LTS4, CH-1015 Lausanne, Switzerland. email: elif.vural@epfl.ch, pascal.frossard@epfl.ch.}
\thanks{This work has been partly funded by the Swiss National Science Foundation under Grant  $200020\_132772$.}}


\maketitle

\begin{abstract}
Manifold models provide low-dimensional representations that are useful for processing and analyzing data in a transformation-invariant way. In this paper, we study the problem of learning smooth pattern transformation manifolds from image sets that represent observations of geometrically transformed signals. In order to construct a manifold, we build a representative pattern whose transformations accurately fit various input images. We examine two objectives of the manifold building problem, namely, approximation and classification. For the approximation problem, we propose a greedy method that constructs a representative pattern by selecting analytic atoms from a continuous dictionary manifold. We present a DC optimization scheme that is applicable to a wide range of transformation and dictionary models, and demonstrate its application to transformation manifolds generated by the rotation, translation and anisotropic scaling of a reference pattern. Then, we generalize this approach to a setting with multiple transformation manifolds, where each manifold represents a different class of signals. We present an iterative multiple manifold building algorithm such that the classification accuracy is promoted in the learning of the representative patterns. Experimental results suggest that the proposed methods yield high accuracy in the approximation and classification of data compared to some reference methods, while the invariance to geometric transformations is achieved due to the transformation manifold model.

\end{abstract}

\begin{IEEEkeywords}
Manifold learning, pattern transformation manifolds, pattern classification, transformation-invariance, sparse approximations.
\end{IEEEkeywords}

%
\IEEEpeerreviewmaketitle

\section{Introduction}
\label{sec:Intro}

\IEEEPARstart{T}{he} representation of high-dimensional signal sets with signal manifolds has several benefits. Manifold models provide concise and low-dimensional representations that facilitate the treatment of signals. In the case of geometric transformation manifolds, the knowledge of the generating model provides a basis for the registration of signals. Moreover, in a setting where different signal classes are represented with different manifolds, the class label of a query signal can be estimated by comparing its distance to the candidate manifolds.

In this work, we focus on pattern transformation manifolds. A pattern transformation manifold (PTM) represents images that are generated from a reference pattern that undergoes a certain set of geometric transformations. For instance, the images obtained by the rotation and scaling of a reference pattern form a PTM. Given a set of visual data that are assumed to be geometrically transformed observations of a signal, we address the problem of constructing a PTM that represents the data accurately. We assume that the type of the transformations that generate the input images, i.e., the transformation model, is known. However, we do not assume any prior alignment of the input images; i.e., the individual transformation parameters corresponding to the images are to be computed. Under these assumptions, our manifold computing problem is formulated as the construction of a representative pattern, together with the estimation of the transformation parameters approximating the input signals. We consider a PTM model that is generated by smooth geometric transformations. We propose to build the representative pattern as a linear combination of some parametric atoms, \revis{which are waveforms that are adapted to the local structures of signals \cite{1028585}}. The atoms are selected from a continuous dictionary manifold that is formed by the smooth geometric transformations of an analytic mother function. The utilization of smooth and parametric atoms in the pattern construction brings desirable properties such as the smoothness of the PTM, and a parametric approximation of the input data, which is useful for effective description of data information. We study the PTM building problem in two parts, where we respectively address approximation and classification applications.

In the data approximation part, we build on our previous work \cite{Vural2011learn} and aim at obtaining an accurate transformation-invariant approximation of input images with the learned manifold.  We iteratively construct a representative pattern by successive addition of atoms such that the total squared distance between the input images and the transformation manifold is minimized. The selection of an atom is then formulated as an optimization problem with respect to the parameters and the coefficient of the atom. We propose a two-stage solution for the atom selection, where we first estimate the parameters of a good atom and then improve this solution. In the first stage, we derive an approximation of the objective function (total squared distance) in a DC (Difference-of-Convex) form; \revis{i.e., in the form of a difference of two convex functions}. We describe a procedure for computing this DC decomposition when a DC form of the geometrically transformed atom is known. The resulting DC approximation is minimized using a DC solver.  Then, we refine the solution of the first stage with a gradient descent method where we approximate the manifold distance by the tangent distance in the objective function. \revis{Although our methodology in this paper is based on ideas very similar to those of \cite{Vural2011learn}, we generalize the setting to arbitrary transformation manifolds, dictionary models and mother functions. In the derivation of the DC decomposition of the objective function, we use some results from \cite{efiDC}, which however targets a different problem that is the alignment of a query image with a reference pattern.}

In the second part of our work, we extend this manifold building approach in order to explore transformation-invariant classification. We consider multiple sets of geometrically transformed observations, where each set consists of a different class of images. We study the problem of constructing multiple PTMs such that each PTM represents one image class, and the images can be accurately classified with respect to their distances to the constructed PTMs. We propose an iterative method that jointly selects atoms for the representative patterns of all classes. We define an objective function that is a weighted combination of a classification and a data approximation error term. Then, we select atoms by minimizing a two-stage approximation of the objective function as in the first part. Experimental results indicate that the approaches proposed for single and multiple manifold computation perform well in transformation-invariant approximation and classification applications in comparison with baseline methods.

The rest of the paper is organized as follows. In Section \ref{sec:relWork} we give a review of related work. In Section \ref{sec:AppxBasedMB}, we discuss the manifold computation problem for transformation-invariant approximation of image signals. Then, in Section \ref{sec:ClassBasedMB}, we present an extension of the proposed scheme for transformation-invariant classification. We discuss the complexity of the proposed methods in Section \ref{sec:complexity}. Finally, we conclude in Section \ref{sec:Conclusion}.

\section{Related Work}
\label{sec:relWork}

Our study is linked to two main topics; manifold learning and sparse signal representations. Firstly, our PTM building approach can be seen as a special instance of manifold learning with prior information on the data model. Manifold learning refers to the recovery of low-dimensional structures in high-dimensional data. Many methods have recently been proposed in this field. The ISOMAP algorithm \cite{266187} computes a global parameterization of data based on the preservation of geodesic distances, while the LLE algorithm \cite{Roweis00nonlineardimensionality} maps the data to a lower-dimensional domain using its locally linear structure. The Hessian Eigenmaps  \cite{donoho03hessian} algorithm has achieved some improvements on LLE, as it also involves some higher-order geometric characteristics of the data. However, such approaches have the following three main shortcomings. First, they compute a  parameterization for the initially available data, and their generalization for the parameterization of additional data is not straightforward. A method has been proposed in \cite{Bengio04out} that provides out-of-sample extensions for some common manifold learning algorithms. The authors interpret these algorithms as learning the eigenvectors of a data-dependent kernel, and then generalize the eigenvectors to the continuous domain in order to compute eigenfunctions. Second, the aforementioned methods lack the means for synthesizing new samples that conform to the same manifold model. This observation is one of the motivations of the method presented in \cite{DollarRabaudBelongieICML07manifold}, which computes a smooth tangent field with the use of analytic functions and thus yields a smooth manifold structure that makes the generation of novel points possible. Also, in \cite{AlvarezMezaVDAC11} a method is proposed for synthesizing new images based on the LLE algorithm. Third, most of the methods that do not allow the synthesis of new data do not have immediate generalizations for classification applications. An exception is the work presented in \cite{deridder:supervised}. The authors propose the SLLE algorithm, where LLE is modified such that the discrimination between different class samples is encouraged in the computation of the data embedding.

\revis{All of the methods mentioned above are generic methods that make no assumption on the type of the manifold underlying the observed data. If they are applied on a data set sampled from a transformation manifold, the embedding computed with these generic methods does not necessarily reflect the real transformation parameters. Our learning algorithm differs from these methods essentially in the fact that it uses the information about the model generating the data, employs it for learning an accurate representation, and also computes the exact transformation parameter vectors.} Since the manifold is constructed in a parametric form,  the mapping between the parameter domain and the high-dimensional signal domain is perfectly known. Thus, one can generate new samples on the manifold and compute the parametrizations of initially unavailable data simply by finding their projections on the manifold. This also permits the estimation of the distance between a test image and the computed manifolds. Consequently, it is possible to assign class labels to test images in a transformation-invariant way by comparing their distances to the computed class-representative manifolds. \revis{Finally, as demonstrated by some of our experiments, the incorporation of the model knowledge into the manifold learning procedure brings important advantages such as robustness to data noise and sparse sampling of data, in comparison with generic methods based on local linearity assumptions.}

\revisthree{The method proposed in \cite{Miller2006} is related to our work in the sense that it computes a simultaneous alignment of a set of images that have undergone transformations, where the application of the method to classification problems is also demonstrated. However, their technique is essentially different from ours as it is based on the idea of ``congealing'' via the minimization of entropy in the corresponding pixels of aligned images.} Next, our paper uses the idea of learning by fitting a parametric model to the data. It is possible to find several other examples of this kind of approach in the literature. For example, the article \cite{atkeson_locally1} is a survey on locally weighted learning, where regression methods for computing linear and nonlinear parametric models are discussed. The efficient computation of locally weighted polynomial regression is the focus of \cite{MooreSD97}. Meanwhile, the method in \cite{Jonsson2007} applies locally weighted regression techniques to the appearance-based pose estimation problem. Then, we remark the following about the relation between this work and the field of sparse signal approximations. Since we achieve a greedy construction of representative patterns, our method bears some resemblance to sparse approximation algorithms such as Matching Pursuit (MP) \cite{1028585} or Simultaneous Orthogonal Matching Pursuit (SOMP) \cite{1140735}. \revis{There are also common points between our method and the Supervised Atom Selection (SAS) algorithm proposed in \cite{EPFLARTICLE91054}, which is a classification-driven sparse approximation method. SAS selects a subset of atoms from a discrete dictionary by minimizing a cost function involving a class separability term and an approximation term.} However, the main contributions of this work in comparison with such algorithms lie in the following. Firstly, we achieve a transformation-invariant approximation of signals due to the transformation manifold model. Furthermore, we employ an optimization procedure for computing the atom parameters that provide an accurate approximation (or classification) of signals. This corresponds to learning atoms from a dictionary manifold, whereas methods such as MP and SOMP pick atoms from a predefined discrete dictionary. This also suggests that it is possible to find connections between our work and transformation-invariant dictionary learning, where a sparse representation of signals is sought not only in terms of the original atoms but also in their geometrically transformed versions. So far, transformation-invariance in sparse approximations has been mostly studied for shift-invariance as in \cite{Mailhe08shift} and \cite{Jost2006_1450}, and for scale-invariance as in \cite{Mairal2007}, \cite{SallOls2002}. The work presented in \cite{Ekanad2011}  also achieves shift-invariance in the sparse decomposition via a continuous basis pursuit. Our new PTM learning method involves the formation of atoms that ensure invariance to a relatively wide range of geometric transformations in comparison with the above works. Our study may thus provide some insight into transformation-invariance in sparse  approximations as well.

%
%
%
\revis{\section{Computation of PTMs for Signal Approximation}}
\label{sec:AppxBasedMB}

\subsection{Problem Formulation}
\label{ssec:AppxProblemForm}

\begin{figure}[t]
 \centering
  \includegraphics[scale=0.4]{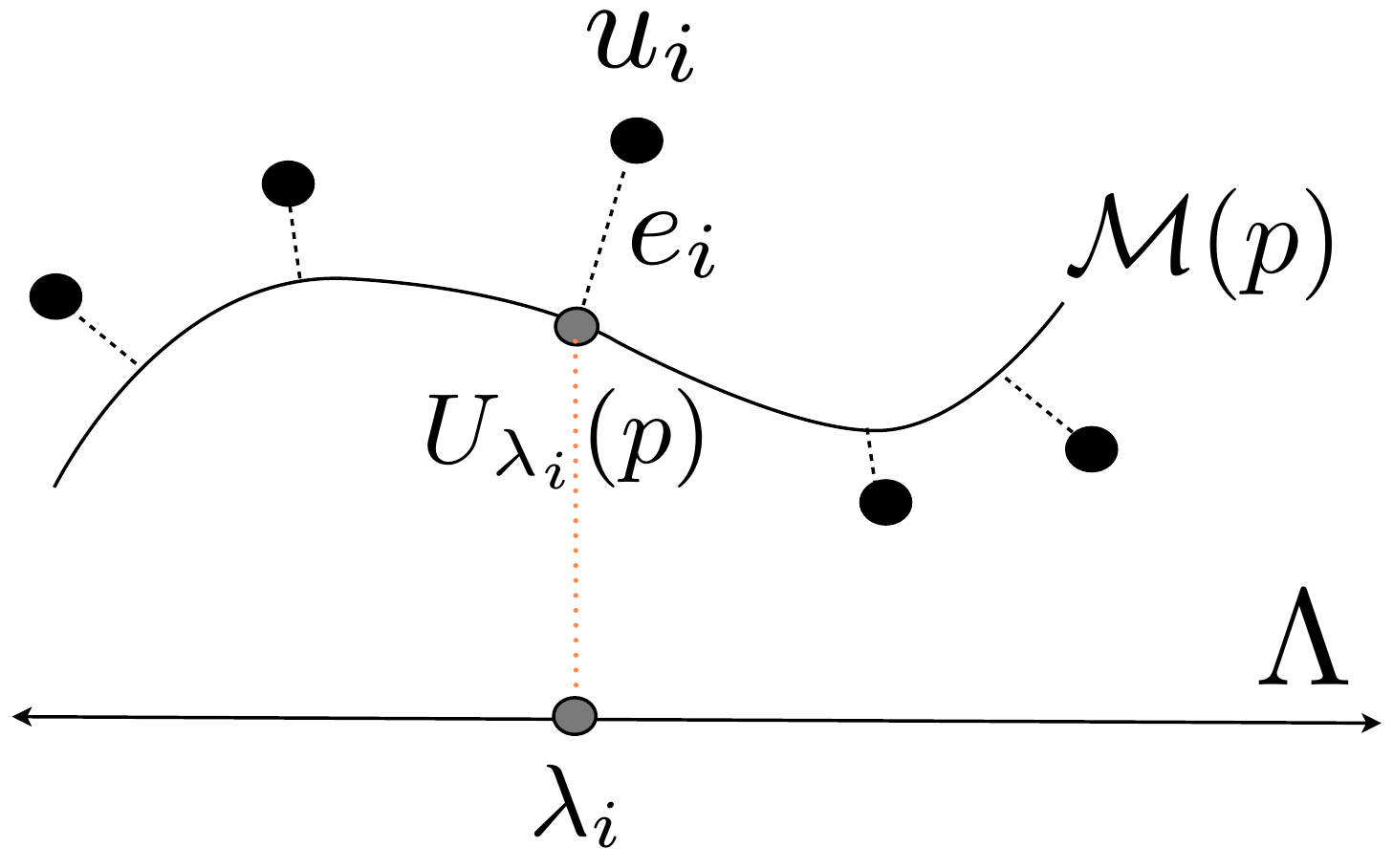}
  \caption{The set $ \{ u_i \} $ of geometrically transformed observations is approximated with the transformation manifold $\mathcal{M}(p)$ of a representative pattern $p$.}
  \label{fig:illusPTMModel}
\end{figure}

\revis{The PTM computation problem can be briefly explained as follows. Given a set of observations $\{ u_i \}$, we would like to compute a  pattern $p$ such that its transformation manifold $\mathcal{M}(p)$ (the set of geometrically transformed versions of $p$) fits the observations $\{ u_i \}$. Therefore, we look for a pattern $p$ such that the total distance between $\mathcal{M}(p)$ and $\{ u_i \}$ is minimized, which is illustrated in Figure \ref{fig:illusPTMModel}. Now we define the problem formally.}
 
Let $p \in L^2(\mathbb{R}^2)$ be a visual pattern, \revis{where $L^2(\mathbb{R}^2)$ denotes the set of square-integrable functions on $\mathbb{R}^2$.} Let  $\Lambda \subset \mathbb{R}^d$ be a closed parameter domain, and $\lambda \in \Lambda$ be a parameter vector. We define $A_{\lambda}(p)  \in L^2(\mathbb{R}^2)$ as the pattern that is generated by applying the geometric transformation specified by $\lambda$ to $p$. \revis{For instance, if $\lambda=(t_x, t_y) $ represents a 2-D translation, then $A_{\lambda}(p)$ corresponds to a translated version of $p$ by $(t_x, t_y)$.} The relation between the two patterns is expressed as $A_{\lambda}(p) (x,y)=p(x', y')$, where the two pairs of coordinate variables are related as $(x',y')=a(\lambda,x,y)$. We assume that $a$ is a smooth ($C^{\infty}$) function. Also, defining $a_\lambda(x,y):=a(\lambda,x,y)$ for a fixed $\lambda \in \Lambda$, we assume that $a_{\lambda}:\mathbb{R}^2 \rightarrow  \mathbb{R}^2$ is a bijection.
%
Then, we define the transformation manifold of $p$ as\footnote{\revistwo{$\mathcal{M}(p)$ is a Riemannian manifold with the Riemannian metric given by $g_{ij} (\lambda) = \langle \frac{\partial U_{\lambda} (p)}{\partial \lambda_i} ,   \frac{\partial U_{\lambda} (p)}{\partial \lambda_j} \rangle$, where $\langle  ,  \rangle$ denotes the usual inner product in $\mathbb{R}^n$ and $\lambda_i$, $\lambda_j$ denote the $i$-th and $j$-th transformation parameters.}}\textsuperscript{,}\footnote{\revistwo{In this paper, we demonstrate our method on the transformation models (\ref{eq:manifoldModelSpecial}) and (\ref{eq:manifoldModelBact}). The transformation manifold of the model in (\ref{eq:manifoldModelBact}) is a transformation group called the similitude group, where the manifold $\mathcal{A}(p)=\{A_{\lambda}(p): \lambda \in \Lambda \} \subset  L^2(\mathbb{R}^2)$ (the counterpart of  $\mathcal{M}(p)$ in the continuous space) corresponds to the group orbit of $p$. However, it should be noted that the model in (\ref{eq:manifoldModelSpecial}) does not correspond to a transformation group.}}
\begin{equation}
\mathcal{M}(p)=\{ U_{\lambda}(p): \lambda \in \Lambda \} \subset \mathbb{R}^n,
\label{eq:manifoldModel}
\end{equation}
%
where $U_{\lambda} (p) \in \mathbb{R}^n$ is an $n$-dimensional discretization of $A_{\lambda}(p)$.\footnote{\revis{When sampling $A_{\lambda}(p)$ to get $U_{\lambda} (p)$, we fix  a rectangular window on $\mathbb{R}^2$, and a regular sampling grid once and for all. Note that defining the pattern transformations in the continuous space $L^2(\mathbb{R}^2)$ instead of $\mathbb{R}^n$, together with constructing $p$ with parametric atoms in $L^2(\mathbb{R}^2)$, saves us from resampling and interpolation ambiguities. In other words, since the sampling grid is fixed and $p(x,y)$, and thus $A_{\lambda}(p)(x,y)$ are analytically known functions, there is a unique way of generating $U_{\lambda} (p)$ for any $\lambda$. Note also that, as explained in Section \ref{ssec:appxPTMGeneric}, our method does not require the transformation of the discrete input images $\{u_i\} \subset \mathbb{R}^n$, i.e., $\{u_i\}$ are used as given, and transformations are always applied to $p$ throughout the algorithm. A rectangular window for sampling is a suitable choice in our application, as the atoms typically have good time-localization (such as Gaussians), and so does $p$. Moreover, if the generation of atoms involves a scaling and translation of the mother function, which is often the case, the method has a natural adaptivity for different window sizes, window locations and sampling rates due to the scalability and position of the atoms, and therefore $p$.}}

\revis{Let $\| . \|$ denote the $\ell_2$-norm in $\mathbb{R}^n$. For a given $u \in  \mathbb{R}^n$, let $\lambda= \arg \min_{\bar{\lambda}} \| u - U_{\bar{\lambda}}(p) \|$. Then $U_{\lambda}(p)$ is called a projection of $u$ on $\mathcal{M}(p)$. In this case, the distance between $u $ and $\mathcal{M}(p)$ is given by\begin{equation}
d(u,\mathcal{M}(p)) = \| u-  U_{\lambda}(p) \| = \min_{\bar{\lambda}\in \Lambda} \| u - U_{\bar{\lambda}}(p) \|.
\end{equation}}

Let $\mathcal{U} = \{ u_i \}_{i=1}^N \subset \mathbb{R}^n$ be a set of observations of a geometrically transformed visual signal. We would like to describe these observations as $ u_i=U_{\lambda_i}(p)+e_i$ by the transformations $U_{\lambda_i}(p)$ of a common representative pattern $p$, where the term $e_i$ indicates the deviation of $u_i$ from $\mathcal{M}(p)$. In the selection of $p$, the objective is to approximate the images in $\mathcal{U}$ accurately. We represent the approximation accuracy in terms of the distance of the input images to $\mathcal{M}(p)$. We formalize this problem as follows. \\

\begin{problem}
\label{prob:AppxGen}
Given images $\mathcal{U} = \{ u_i \}_{i=1}^N $, compute a pattern $p \in L^2(\mathbb{R}^2)$ and a set of transformation parameter vectors $\{\lambda_i\}_{i=1}^N \subset \Lambda$, by minimizing

\begin{equation}
\label{eq:probdefnEgen}
E=\sum_{i=1}^N  \| u_i - U_{\lambda_i} (p)  \|^2.
\end{equation}
\end{problem}

The error $E$ corresponds to the total squared distance of the input images to $\mathcal{M}(p)$. In order to solve Problem \ref{prob:AppxGen}, we propose to construct $p$ as a sparse linear combination of some parametric atoms from a dictionary manifold

\begin{equation}
\label{eq:dictManifold}
\mathcal{D}  = \{  B_{\gamma}(\phi): \gamma  \in \Gamma  \} \subset L^2(\mathbb{R}^2).
\end{equation}
%
Here, each atom $B_{\gamma}(\phi) \in  L^2(\mathbb{R}^2)$ is derived from the analytic mother function $\phi \in L^2(\mathbb{R}^2)$ through a geometric transformation specified by a parameter vector $\gamma$. An atom is thus given by $B_{\gamma}(\phi) (x,y)=\phi(x', y')$, where $(x',y')=b(\gamma,x,y)$. We assume that $b$ is a smooth function, and that $b_{\gamma}(x, y): =b(\gamma, x, y)$, $b_{\gamma}:\mathbb{R}^2 \rightarrow  \mathbb{R}^2$ is a bijection for any fixed $\gamma \in \Gamma$. The parameter domain $\Gamma$ is assumed to be a closed and convex subset of $\mathbb{R}^s$ for some $s$, where $s$ is the number of transformation parameters generating $\mathcal{D}$. Hence, $\mathcal{D}$ is an $s$-manifold. Let us write $\phi_{\gamma} = B_{\gamma}(\phi) $ for simplicity.  We would like to obtain the representative pattern in the form $p=\sum_{j=1}^K c_j  \, \phi_{\gamma_j} $ as a combination of $K$ atoms $\{ \phi_{ \gamma_j}\}$ with coefficients $\{c_j\}$. Under these assumptions, we reformulate the previous problem as follows.\footnote{\revis{Whether the span of the dictionary $\mathcal{D}$ is dense in $L^2(\mathbb{R}^2)$ depends on the mother function $\phi$ as well as the transformation $b$. In this paper, we present results where $\mathcal{D}$ is generated by the rotation, anisotropic scaling and translation of the mother function. The proof of Proposition 2.1.2 in \cite{Antoine2004} shows that for this very transformation model, the linear span of $\mathcal{D}$ is dense in $L^2(\mathbb{R}^2)$ as long as $\phi$ has nontrivial support, i.e., unless $\phi(x,y)=0$ almost everywhere.}}\\

\begin{problem}
\label{prob:AppxDict}
Given images $\mathcal{U} = \{ u_i \}_{i=1}^N $, an analytic mother function $\phi$, and a sparsity constraint $K$; compute a set of atom parameter vectors $\{ \gamma_j \}_{j=1}^K \subset \Gamma$, a set of coefficients $\{c_j\}_{j=1}^K \subset \mathbb{R}$, and a set of transformation parameter vectors $\{\lambda_i\}_{i=1}^N \subset \Lambda$, by minimizing

\begin{equation}
\label{eq:probdefnEdict}
E=\sum_{i=1}^N  \| u_i - U_{\lambda_i} \big( \sum_{j=1}^K c_j \,  \phi_{\gamma_j}  \big)  \|^2.
\end{equation}
\end{problem}

Note that the construction of $p$ with smooth atoms assures the smoothness of the resulting transformation manifold. A manifold point $U_{\lambda}(p) \in \mathbb{R}^n$ is given by the discretization of the function
\begin{eqnarray}
\begin{split}
 A_{\lambda}(p)(x,y) &= p(a_{\lambda}(x,y)) = \sum_{j=1}^K c_j \, \phi_{\gamma_j}(a_{\lambda} (x,y) )\\
 & =  \sum_{j=1}^K c_j \, \phi( b_{\gamma_j} \circ a_{\lambda} (x,y) )
\end{split}
\end{eqnarray}
\revis{where the notation $\circ$ stands for function composition}. Here $a_{\lambda}(x,y)$ is a smooth function of $\lambda$; and $b$ and $\phi$ are smooth functions, too. Therefore, $A_{\lambda}(p)(x,y)$ is a smooth function of $\lambda$. Then, each component $U_{\lambda}(p) (l)$ of $U_{\lambda}(p)$ is a smooth function of $\lambda$, for $l=1, \, \dots \,, n$.

%
%
%

\subsection{PTM Building Algorithm}
\label{ssec:appxPTMGeneric}

We now build on our previous work \cite{Vural2011learn} and describe an algorithm for the solution of Problem \ref{prob:AppxDict}. Due to the complicated dependence of $E$ on the atom and projection parameters, it is hard to find an optimal solution for Problem \ref{prob:AppxDict}. Thus, we propose here a constructive approach. We build the pattern $p$ iteratively by selecting atoms from $\mathcal{D}$ in a greedy manner. Each successive version $p_j$ of the pattern $p$ leads to a different manifold $\mathcal{M}(p_j)$, whose form gradually converges to the the final solution $\mathcal{M}(p)$. \revis{During the optimization of the atom parameters in each iteration, we first locate a good initial solution by minimizing a DC approximation of the objective function using DC programming. We then refine our solution by using a locally linear approximation of the manifold near each input image and minimizing the total tangent distance to the manifold with gradient descent. The reason for our choice of a two-step optimization in atom selection is the following. The DC solver used in our implementation is the cutting plane algorithm, which slows down as the number of vertices increases throughout the iterations. Therefore, in practice, we use the DC programming step for approaching the vicinity of a good solution and we terminate it when it slows down. Then, we continue the minimization of the function with gradient descent. Considering that  the DC program is not affected by local minima and gradient descent is susceptible to local minima, using these two methods respectively for the first and second parts is a suitable choice.} We start by giving a brief discussion of DC functions \cite{introGlobOpt} that are used in our algorithm.\\

\begin{definition}
A real valued function $f$ defined on a convex set $C \subset \mathbb{R}^s$ is called DC on $C$ if for all $x \in C$, $f$ can be expressed in the form 
\begin{equation}
\label{eq:dcdefn}
f(x) = g(x) - h(x)
\end{equation}
where $g$, $h$ are convex functions on $C$. The representation (\ref{eq:dcdefn}) is said to be a DC decomposition of $f$.\\
\end{definition}

An important fact about DC functions is the following\footnote{Proposition \ref{prop:smoothIsDC} is the original statement of Corollary 4.1 in \cite{introGlobOpt}, which holds for functions defined on $\mathbb{R}^s$. However, a function defined on a convex subset of $\mathbb{R}^s$ with continuous second partial derivatives is also DC. This can be easily seen by referring to the proof of Corollary 4.1 in \cite{introGlobOpt}, which is based on the fact that locally DC functions are DC, and to Hartman's proof \cite{Hartman1959} that locally DC functions defined on a convex subset of $\mathbb{R}^s$ are DC on the same domain.} \cite{introGlobOpt}.\\

\begin{proposition}
\label{prop:smoothIsDC}
Every function $f: \mathbb{R}^s \rightarrow \mathbb{R}$ whose second partial derivatives are continuous everywhere is DC.\\
\end{proposition}
 
 \begin{figure}[t]
 \centering
  \includegraphics[scale=0.4]{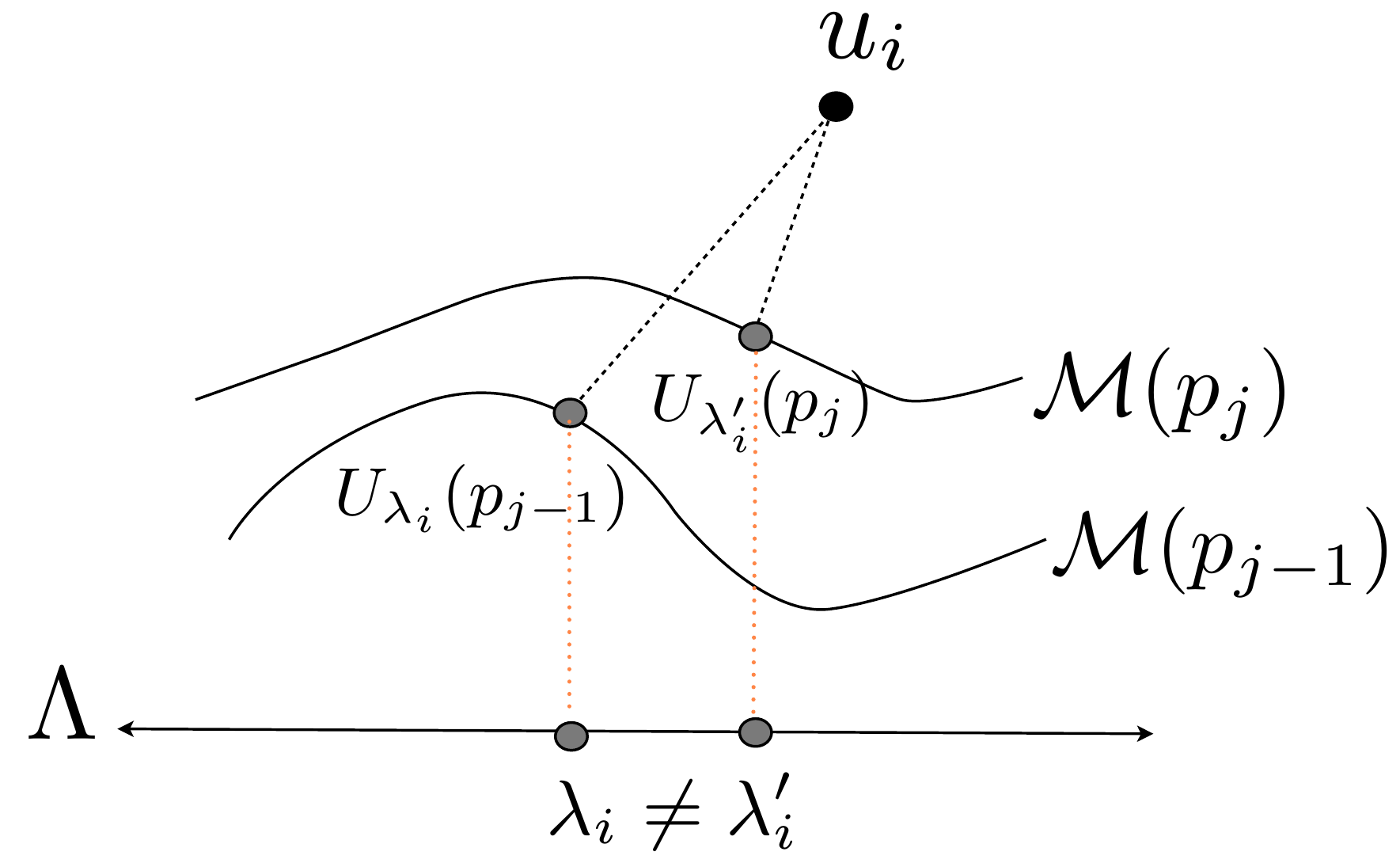}
  \caption{The parameter vectors corresponding to the projections of the point $u_i$ on the previous manifold $\mathcal{M}(p_{j-1})$ and the updated manifold $\mathcal{M}(p_j)$ are shown respectively by $\lambda_i$ and $\lambda'_i$.}
  \label{fig:illusProjUpdate}
\end{figure}

\begin{figure}[t]
 \centering
  \includegraphics[scale=0.4]{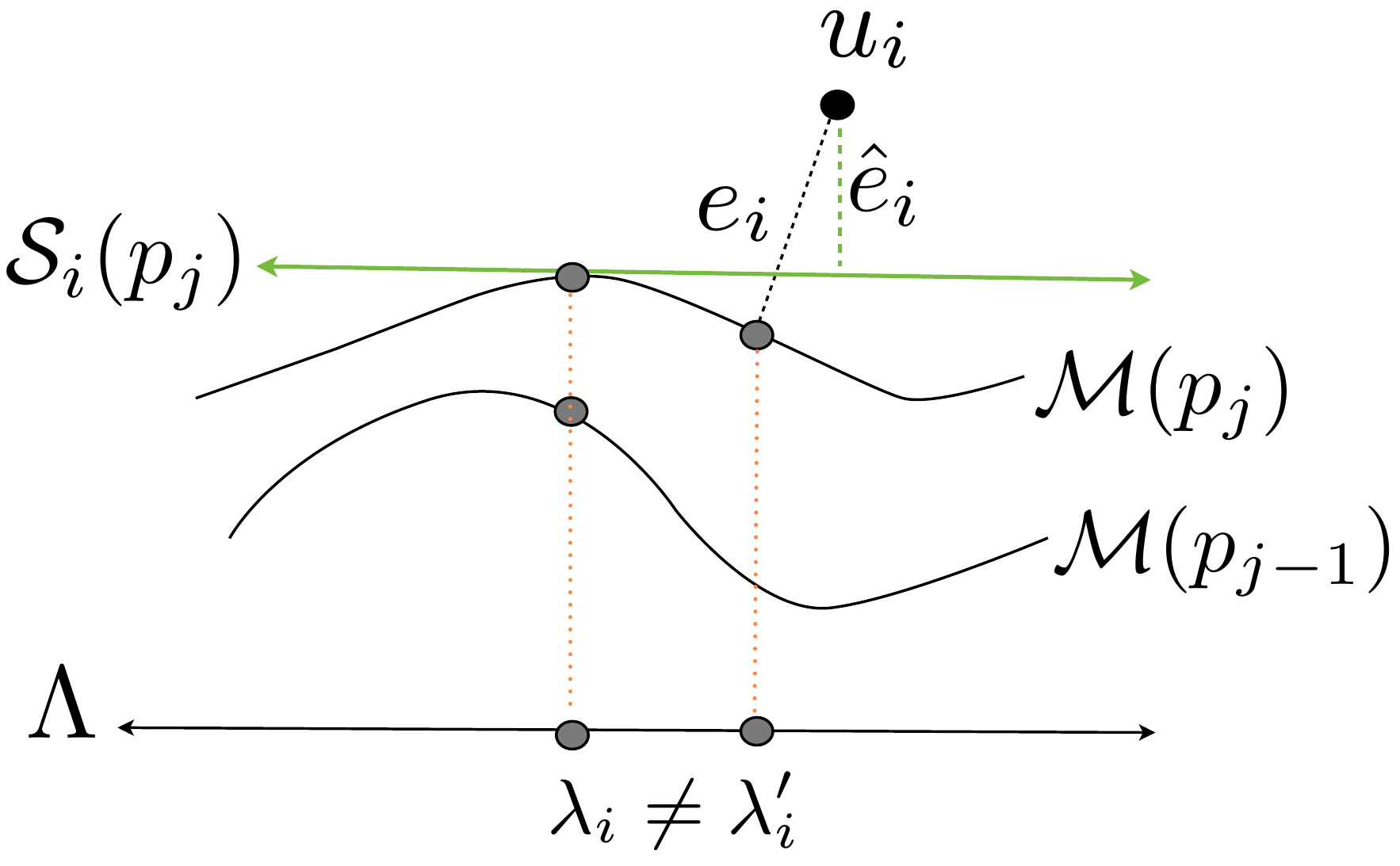}
  \caption{$\mathcal{S}_i(p_j)$ is the first order approximation of the manifold $\mathcal{M}(p_j)$ around $U_{\lambda_i}(p_j)$. Here, the difference vector $e_i$ between $u_i$ and its exact projection on $\mathcal{M}(p_j)$ is approximated by the difference vector $\hat{e}_i$ between $u_i$ and its projection on $\mathcal{S}_i(p_j)$.}
  \label{fig:illusTangDist}
\end{figure}
 
The global minimum of DC functions can be computed using DC solvers such as the cutting plane algorithm and the branch-and-bound algorithm \cite{Horst1999}, which is a major reason for the choice of DC programming in this work. There are also some DC optimization methods such as DCA \cite{Tao97} and the concave-convex procedure (CCCP) \cite{Yuille02theconcave}, which have favorable computational complexities and converge to a local minimum. The theoretical guarantee for finding the global minimum with the cutting plane algorithm is lost when the DC program is terminated before exact convergence as in our implementation; however, the overall two-step minimization  gives good results in practice.

Equipped with the DC formalism, we can now describe our iterative manifold learning algorithm. As the atom selection procedure requires the computation of the distance between the input images and the PTM, the algorithm initially needs a rough estimate of the parameter vectors. Therefore, we first assign a tentative set of parameter vectors $\{ \lambda_i\}$ to the images $\{u_i\}$ by projecting $\{u_i\}$ onto some reference transformation manifold $\mathcal{M}(\Psi)$. The pattern $\Psi$ can be possibly chosen as a typical pattern in the input set (an $L^2(\mathbb{R}^2)$-representation of some $u_i$). Then, the parameter vector assigned to an image is given by $ \lambda_i = \arg \min_{\lambda \in \Lambda} \| u_i - U_{\lambda}(\Psi)  \| $. \revis{We compute the transformation parameters by first roughly locating the projections with the help of a grid, and then performing a line search near the closest grid point.}

Now let us describe the $j$-th iteration of the algorithm. Let $p_{j-1}$ denote the pattern consisting of $j-1$ atoms (one can set $p_0$=0). In the $j$-th iteration we would like to choose an atom $\phi_{\gamma_j} \in \mathcal{D}$ and a coefficient $c_j $ such that the data approximation error 
\begin{equation}
\label{eq:appErrorIter}
E=\sum_{i=1}^N \| e_i \|^2 =\sum_{i=1}^N  d^2 ( u_i,  \mathcal{M}( p_j )  )
\end{equation}
is minimized, where $p_j=p_{j-1}+c_j \, \phi_{\gamma_j} $. We remark that the cost function in (\ref{eq:probdefnEdict}) is defined as a function of all atom parameters $\{\gamma_j\}_{j=1}^K$ and coefficients $\{c_j\}_{j=1}^K$, however, the one in (\ref{eq:appErrorIter}) is considered only as a function of $\gamma_j$ and $c_j$. For simplicity, we use the same symbol $E$ for these two functions with an abuse of notation.

Notice that the values of $ \{ \lambda_i \}$ may change between iterations $j-1$ and $j$, because the projection points change when the manifold is updated. The alteration of $ \{ \lambda_i \}$ is illustrated in Figure \ref{fig:illusProjUpdate}. At the beginning of the $j$-th iteration, the vectors $ \{ \lambda_i \}$ take the values computed at the end of iteration $j-1$ by projecting $\{  u_i \}$ on $\mathcal{M}(p_{j-1})$. Therefore, $d(u_i,  \mathcal{M}( p_j ) ) \neq \| u_i - U_{\lambda_i} (p_j) \|  $ in general.  In the minimization of $E$, it is not easy to formulate and compute the exact distance $d(u_i,  \mathcal{M}( p_j ) )$, since it would require the formulation of $  \lambda_i $ as a function of the optimization variables, which does not have a known closed-form expression. Therefore, we propose to minimize $E$ in two stages. Let $\gamma=\gamma_j$ and $c=c_j$ denote the parameters and the coefficient of the new atom for the ease of notation. In the first stage, we define a coarse approximation\footnote{\revis{The operator $U_{\lambda}$ is linear, since for two patterns $p$, $r$, and a scalar $c$, we have $A_{\lambda}(c \, p +r) (x,y)=(c \, p +r)(a_{\lambda}(x,y)) = c \, p(a_{\lambda}(x,y)) + r(a_{\lambda}(x,y)) = c A_{\lambda}(p)(x,y) + A_{\lambda}(r)(x,y)$.}} 
\begin{eqnarray}
\begin{split}
\tilde{E}& = \sum_{i=1}^N  \| \tilde{e}_i \|^2 = \sum_{i=1}^N  \| u_i - U_{\lambda_i} (  p_{j-1}  +   c \,\phi_{\gamma} )  \|^2 \\
&= \sum_{i=1}^N  \| v_i -  c \, U_{\lambda_i} (  \phi_{\gamma} )  \|^2 
\end{split}
\label{eq:tildeE}
\end{eqnarray}
of $E$, where $v_i=u_i - U_{\lambda_i} ( p_{j-1})$ is a constant with respect to $\gamma$ and $c$.  We have the following proposition.\\

\begin{proposition}
\label{prop:DCformTildeE}
$\tilde{E}$ is a DC function of $\gamma$ and $c$. Moreover, if a DC decomposition for the components (pixels) of the transformed atom $U_{\lambda} (  \phi_{\gamma} )$ is known, a DC decomposition of $\tilde{E}$ is computable.\\
\end{proposition}

The proof of Proposition \ref{prop:DCformTildeE} is given in Appendix A. \revis{Although finding the DC decomposition of an arbitrary function is an open problem, DC decompositions are available for important function classes \cite{Horst1999}. See, for instance, \cite{efiDC} for the derivation of the DC decompositions of several elementary functions, and \cite{Horst1999} for operations with known DC decompositions.} For the rest of our discussion, we assume that a DC decomposition of the components of $U_{\lambda_i} (  \phi_{\gamma} )$ is computable. We can therefore minimize $\tilde{E}$ using the cutting plane algorithm discussed in \cite{Horst1999} and \cite{efiDC}. This provides an initial solution for the atom that is optimized further in the next stage.

In the second stage of our method, we approximate $E$ by another function $\hat{E}$, which is the sum of the squared tangent distances of $\{ u_i \}$ to the updated manifold $\mathcal{M}(p_j)$. Let $\mathcal{S}_i (p_j)$ denote the first order approximation of  $\mathcal{M}(p_j)$ around $U_{\lambda_i}(p_j)$, where $\lambda_i$ is still as computed at the end of iteration $j-1$. Then, the distance $d(u_i, \mathcal{S}_i(p_j ))$ between $u_i$ and $\mathcal{S}_i (p_j)$ is called the tangent distance \cite{668381} and it provides an approximation for $d( u_i,  \mathcal{M}( p_j))$  (illustrated in Figure \ref{fig:illusTangDist}). Hence, $\hat{E}$ is given by

\begin{equation}
\label{eq:hatEdefn}
\hat{E} = \sum_{i=1}^N \| \hat{e}_i \|^2 = \sum_{i=1}^N d^2(u_i, \mathcal{S}_i(p_j)) .
\end{equation} 
The complete derivations of $\hat{E}$, $\mathcal{S}_i(p_j)$ and the distance to $\mathcal{S}_i(p_j)$ are given in Appendix B, where we use results from our previous work \cite{Vural2011appx}. We minimize $\hat{E}$ over $({\gamma}, c)$ using a gradient descent algorithm. At the end of this second stage, we finally obtain our solution for the atom parameters $\gamma$ and the coefficient $c$.

The new atom is then added to the representative pattern such that  $p_j=p_{j-1}+c \,  \phi_{\gamma}$. Since $p_j$ is updated, we recompute the projections of $\{u_i\}$ on the new manifold $\mathcal{M}(p_j)$ and update $\{\lambda_i\}$ such that they correspond to the new projection points. The projections can be recomputed by performing a search in a small region around their previous locations.

\revis{We continue the iterative approximation algorithm until the change in $E$ becomes insignificant or a predefined sparsity constraint is reached. We also finalize the algorithm in case an update increases $E$, which might occur as the atom selection is done by minimizing the approximations of $E$. The termination of the algorithm is guaranteed as $E$ is forced to be non-increasing throughout the iterations. However, due to the complicated structure of the method that uses several approximations of $E$, it is hard to provide a theoretical guarantee that the solution $p$, $\{\lambda_i\}_{i=1}^N$ converges, even if that has been the case in all experiments.} We name this method Parameterized Atom Selection (PATS) and summarize it in Algorithm \ref{alg:pats}. The complexity of the algorithm will be discussed in Section \ref{sec:complexity}. \revis{As a final remark, we discuss the accuracy of reformulating of the objective function $E$ in (\ref{eq:probdefnEgen}) in several stages of the algorithm. Firstly, the error arising from approximating (\ref{eq:probdefnEgen}) with (\ref{eq:probdefnEdict}) asymptotically approaches 0 as the number of atoms in the sparse approximation is increased, provided that the span of $\mathcal{D}$ is dense in  $L^2(\mathbb{R}^2)$. Then, the gradual minimization of (\ref{eq:probdefnEdict}) via minimizing (\ref{eq:appErrorIter}) also introduces an error, which is a common feature of greedy algorithms. Next, the deviation of $\tilde E$ in (\ref{eq:tildeE}) from $E$ in (\ref{eq:appErrorIter}) mainly depends on the amount of change in the  transformation parameters between consecutive iterations. Starting the algorithm with a good initialization of  parameters helps to reduce this error. Moreover, the inaccuracy  caused by this approximation is partially compensated for in the next stage as $\hat E$ accounts for parameter changes. The accuracy of this second approximation essentially depends on the nonlinearity of the manifold; i.e., $\hat E = E$ if the manifold is linear. However, even if the manifold has high curvature the approximation $\hat E \approx E$ is accurate if the change in the transformation parameters is small between adjacent iterations, which is often the case, particularly in the late phases of the algorithm.} \\

\begin{algorithm}[h]
\caption{Parameterized Atom Selection (PATS)}

\begin{algorithmic}[1]

\STATE
\textbf{Input:} \\
$\mathcal{U}=\{ u_i \}_{i=1}^N$: Set of observations\\

\STATE
\textbf{Initialization:}

\STATE
\label{algPATS:lineInit}
Determine a tentative set of parameter vectors $\{ \lambda_i \}$ by projecting $\{ u_i \}$ on the transformation manifold $\mathcal{M}(\Psi)$ of a reference pattern $\Psi$.

\STATE
$p_0=0$.

\STATE
$j=0$.

\REPEAT

\STATE
$j=j+1$.

\STATE
Optimize the parameters ${\gamma}$ and the coefficient $c$ of the new atom with DC programming such that the error $\tilde{E}$ in  (\ref{eq:tildeE}) is minimized.

\STATE
\label{algPATS:lineGD}
Further optimize ${\gamma}$ and $c$ with gradient descent by minimizing the error $\hat E$ in  (\ref{eq:hatEdefn}).

\STATE
Update $p_j=p_{j-1}+c \,  \phi_{\gamma}$.

\STATE
Update parameter vectors $\{ \lambda_i \}$ by projecting $\{ u_i \}$ onto $\mathcal{M}(p_j)$.

\UNTIL the approximation error $E$ converges or increases

\STATE
\textbf{Output}:\\
 $p=p_j$: A representative pattern whose transformation manifold $\mathcal{M}(p)$ fits the input data $\mathcal{U}$\\

\end{algorithmic}
\label{alg:pats}
\end{algorithm}

%
%
%
%

\subsection{Experimental Results}
\label{ssec:appxExperiments}

We now present experimental results demonstrating the application of  PATS in transformation-invariant image approximation. We first describe the experimental setup. We experiment on a PTM model given by 
\begin{equation}
\mathcal{M}(p)=\{ U_{\lambda}(p): \lambda = (\theta, t_x, t_y, s_x, s_y) \in \Lambda \} \subset \mathbb{R}^n
\label{eq:manifoldModelSpecial}
\end{equation}
where $\theta$ denotes a rotation, $t_x$ and $t_y$ represent  translations in $x$ and $y$ directions, and $s_x$ and $s_y$ define an anisotropic scaling in $x$ and $y$ directions. $U_{\lambda} (p)$ is a discretization of $A_{\lambda} (p)$, where $A_{\lambda} (p) (x,y)=p(x', y')$ and
%
\begin{equation}
\label{eq:TransModel}
 \left[
\begin{array}{c}
x' \\
y'
\end{array} \right]
=
\left[
\begin{array}{c c}
 s^{-1}_x & 0  \\
 0 & s^{-1}_y
\end{array} \right]
 \left[
\begin{array}{c c}
 \cos\theta& \sin\theta  \\
 -\sin\theta &\cos\theta
\end{array} \right]
 \left[
\begin{array}{c}
 x - t_x \\
 y - t_y
\end{array} \right]
.
\end{equation}\\
We choose a dictionary manifold model given by
\begin{equation}
\label{eq:dictManifoldExp}
\mathcal{D}  = \{  B_{\gamma}(\phi): \gamma = (\psi, \tau_x, \tau_y, \sigma_x, \sigma_y) \in \Gamma  \} \subset L^2(\mathbb{R}^2)
\end{equation}
where $\psi$ is a rotation parameter, $\tau_x$ and $ \tau_y$ denote translations in $x$ and $y$ directions, and $\sigma_x$ and $ \sigma_y$ represent anisotropic scalings in $x$ and $y$ directions. The geometric transformation between the mother function and an atom is thus given by $\phi_{\gamma}(x,y)=\phi(x', y')$, where
\begin{equation}
\label{eq:DictModel}
 \left[
\begin{array}{c}
x' \\
y'
\end{array} \right]
=
 \left[
\begin{array}{c c}
 \sigma_x^{-1} & 0  \\
 0 & \sigma_y^{-1}
\end{array} \right]
 \left[
\begin{array}{c c}
 \cos \psi& \sin \psi \\
 -\sin \psi &\cos\ \psi
\end{array} \right]
 \left[
\begin{array}{c}
 x - \tau_x  \\
 y - \tau_y
\end{array} \right]
.
\end{equation}

The mother function $\phi$ is taken as a Gaussian function.
\begin{equation}
\label{eq:gaussMotherfunc}
\phi(x,y)=\sqrt \frac{2}{ \pi} e^{-(x^2+y^2)}
\end{equation}
In Appendix C, we describe the computation of the DC decompositions of $U_{\lambda} (\phi_{\gamma})$ and the error $\tilde{E}$ for this  setup.

In the first set of experiments, we test the PATS algorithm on two data sets, which consist of handwritten ``5'' digits and face images. The first data set is generated from the MNIST handwritten digits database \cite{lecun98} by applying random geometric transformations to 30 randomly selected images of the ``5''  digit. The second data set consists of 35 geometrically transformed face images of a single subject with facial expression variations \cite{CKFaceDatabase}, which is regarded as a source of deviation of the data from the manifold. Both data sets are generated by applying rotations, anisotropic scalings and translations.

In the experiments we measure the data approximation error of the learned pattern, which is the average squared distance of input images to the computed transformation manifold. In the plots, the data approximation error is normalized with respect to the average squared norm of input images.

In order to evaluate the performance of the PATS method, we compare it with the following baseline approaches. 
\begin{itemize}
\item 
MP on average pattern: We determine a representative pattern (average pattern) by picking the untransformed image that is closest to the centroid of all untransformed data set images. Then, we obtain progressive approximations of the average pattern with Matching Pursuit \cite{1028585}. 
\item 
SMP on aligned patterns: We obtain a progressive simultaneous approximation of untransformed images with the Simultaneous Matching Pursuit algorithm explained in \cite{SalaLlonch2010}. SMP selects in each iteration one atom that approximates all images simultaneously, but the coefficient of the atom is different for each image. We construct a pattern gradually by adding the atoms chosen by SMP and weighting them with their average coefficients.
%
\item 
Locally linear approximation: We compute the locally linear approximation error, which is the average distance between an image and its projection onto the plane passing through its nearest neighbors. We include this error, since typical manifold learning algorithms such as \cite{Roweis00nonlineardimensionality} and \cite{donoho03hessian} use a linear approximation of the manifold.
\end{itemize}

\begin{figure}[]
\begin{center}
     \subfigure[Images from the digits data set]
       {\label{fig:databaseDigit5PATS}\includegraphics[height=1.6cm]{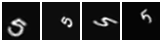}}
     \subfigure[Learned pattern]
       {\label{fig:learnedDigit5PATS}\includegraphics[height=1.5cm]{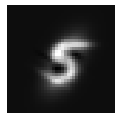}}
      \subfigure[Approximation error]
       {\label{fig:errorDigit5PATS}\includegraphics[width=8cm]{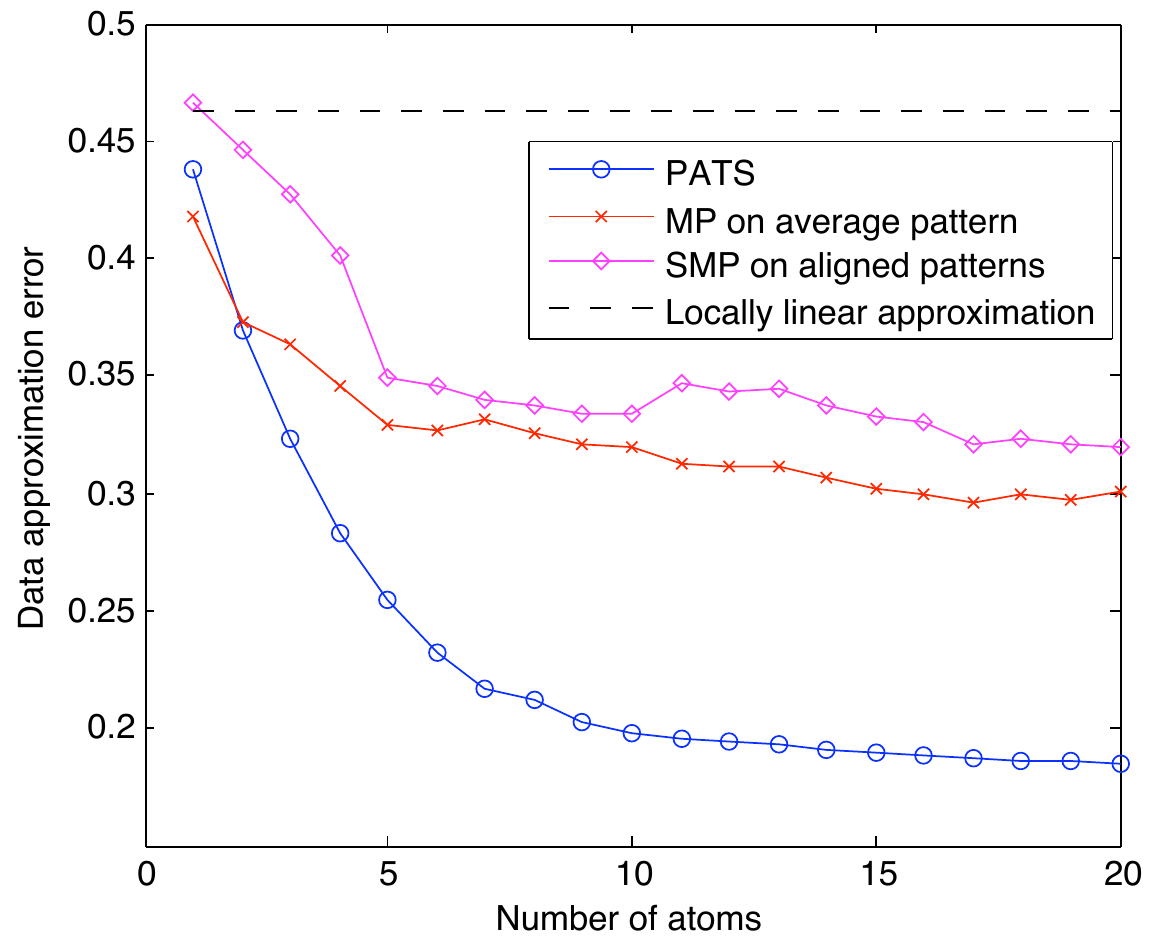}} 
 \end{center}
 \caption{Manifold approximation results with handwritten digits (``5'' )}
 \label{fig:resultsDigit5PATS}
\end{figure}

\begin{figure}[]
\begin{center}
     \subfigure[Images from the face data set]
       {\label{fig:databaseFacePATS}\includegraphics[height=1.6cm]{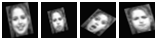}}
     \subfigure[Learned pattern]
       {\label{fig:learnedFacePATS}\includegraphics[height=1.5cm]{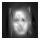}}
      \subfigure[Approximation error]
       {\label{fig:errorFacePATS}\includegraphics[width=8cm]{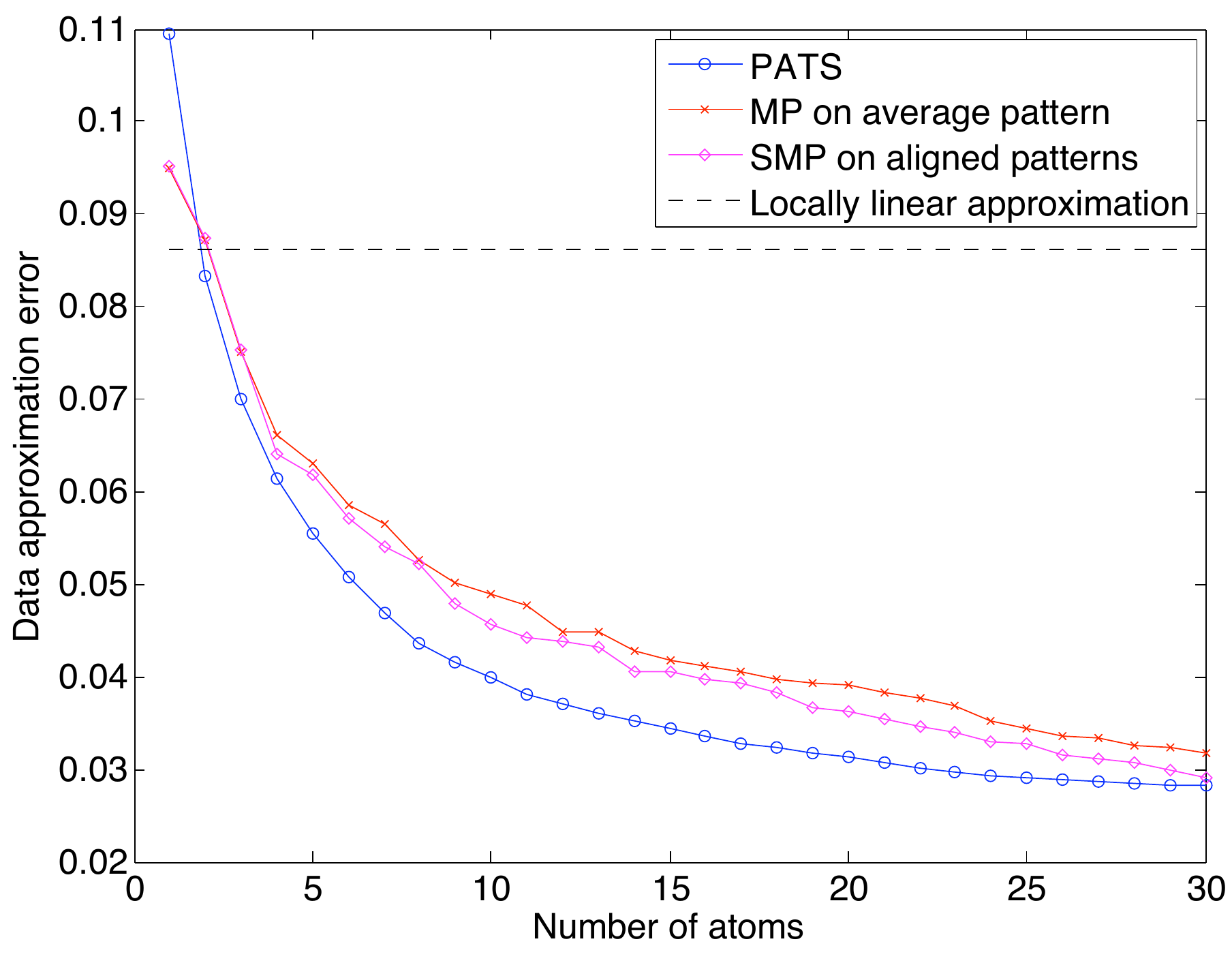}} 
 \end{center}
 \caption{Manifold approximation results with face images}
 \label{fig:resultsFacePATS}
\end{figure}

The dictionary used in the first two methods above is a redundant sampling of the dictionary manifold in (\ref{eq:dictManifoldExp}). The results obtained on the digit and face images are given respectively in Figures \ref{fig:resultsDigit5PATS} and \ref{fig:resultsFacePATS}. Some  images from each data set are shown in Figures \ref{fig:databaseDigit5PATS} and \ref{fig:databaseFacePATS}. The patterns built with the proposed method are displayed in Figures \ref{fig:learnedDigit5PATS} and \ref{fig:learnedFacePATS}. It is seen that the common characteristics of the input images are well captured in the learned patterns. The data approximation errors of the compared methods are plotted in Figures \ref{fig:errorDigit5PATS} and \ref{fig:errorFacePATS}. The errors of the PTM-based methods are plotted with respect to the number of atoms used in the progressive generation of patterns. The results show that the proposed method provides a better approximation accuracy than the other approaches. The approximation accuracies of MP and SMP are better in the face images experiment compared to the digits experiment. This can be explained by the fact that face images of the same subject have smaller numerical variation with respect to handwritten digit images; therefore, an average pattern in the data set can approximate the others relatively well.\footnote{\revis{In an evaluation on the aligned and normalized versions of the input images, the average squared distance to the centroid is found as 0.40 for the digit images and 0.01 for the face images.}} One can also observe that the locally linear approximation error is significantly high. The local linearity assumption fails in these experiments because of the sparse sampling of the data (small number of images), whereas PTM-based methods are much less affected by such sampling conditions.

\revis{In a second experiment, we study the effect of occlusions and outliers in PTM building. We experiment on the same digits data set as before, with a transformation model consisting of 2-D translations, where only the parameters $t_x, t_y$ in (\ref{eq:manifoldModelSpecial}) are used. The images are randomly occluded with horizontal and vertical stripes as shown in Figure \ref{fig:databaseDigitsTrans}. We generate four different data sets, where the first one consists of 150 images of only the digit ``2''. We obtain the other data sets by adding the first data set outliers consisting of a mixture of ``3'', ``5'' and ``8'' digits, where the outlier/inlier ratio is 10\%, 20\% and 30\%. We test the PATS method using a dictionary generated with the inverse multiquadric mother function given by  
\[
\phi(x,y)=(1+x^2+y^2)^{\mu}, \, \, \, \mu<0.
\]
We have set $\mu=-3$ in the experiments.  The computation of the DC decomposition for this mother function is explained in Appendix C. The patterns learned with all four data sets are shown in Figure \ref{fig:learnedDigitsTransMQ}, and the errors are plotted in Figure \ref{fig:errorsDigitTransMQ}. The errors obtained with SMP on aligned patterns are also given for comparison. It is shown that the proposed method can recover a representative ``2'' digit in spite of the occlusions. As the ratio of outliers is augmented, the characteristics of the learned pattern gradually diverge from the ``2'' digit; and the approximation error increases as the average deviation of the data from the ``2'' manifold is increased.

\begin{figure}[]
\begin{center}
     \subfigure[Images from the occluded digits data set]
       {\label{fig:databaseDigitsTrans}\includegraphics[height=1.6cm]{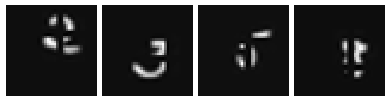}}
       \subfigure[Learned patterns.  From left to right: 0\%, 10\%, 20\%, 30\% outliers.]
       {\label{fig:learnedDigitsTransMQ}\includegraphics[height=1.5cm]{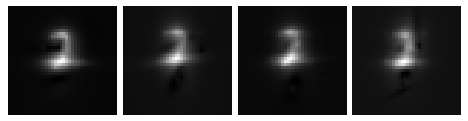}}
       \subfigure[Approximation error]
       {\label{fig:errorsDigitTransMQ}\includegraphics[width=8cm]{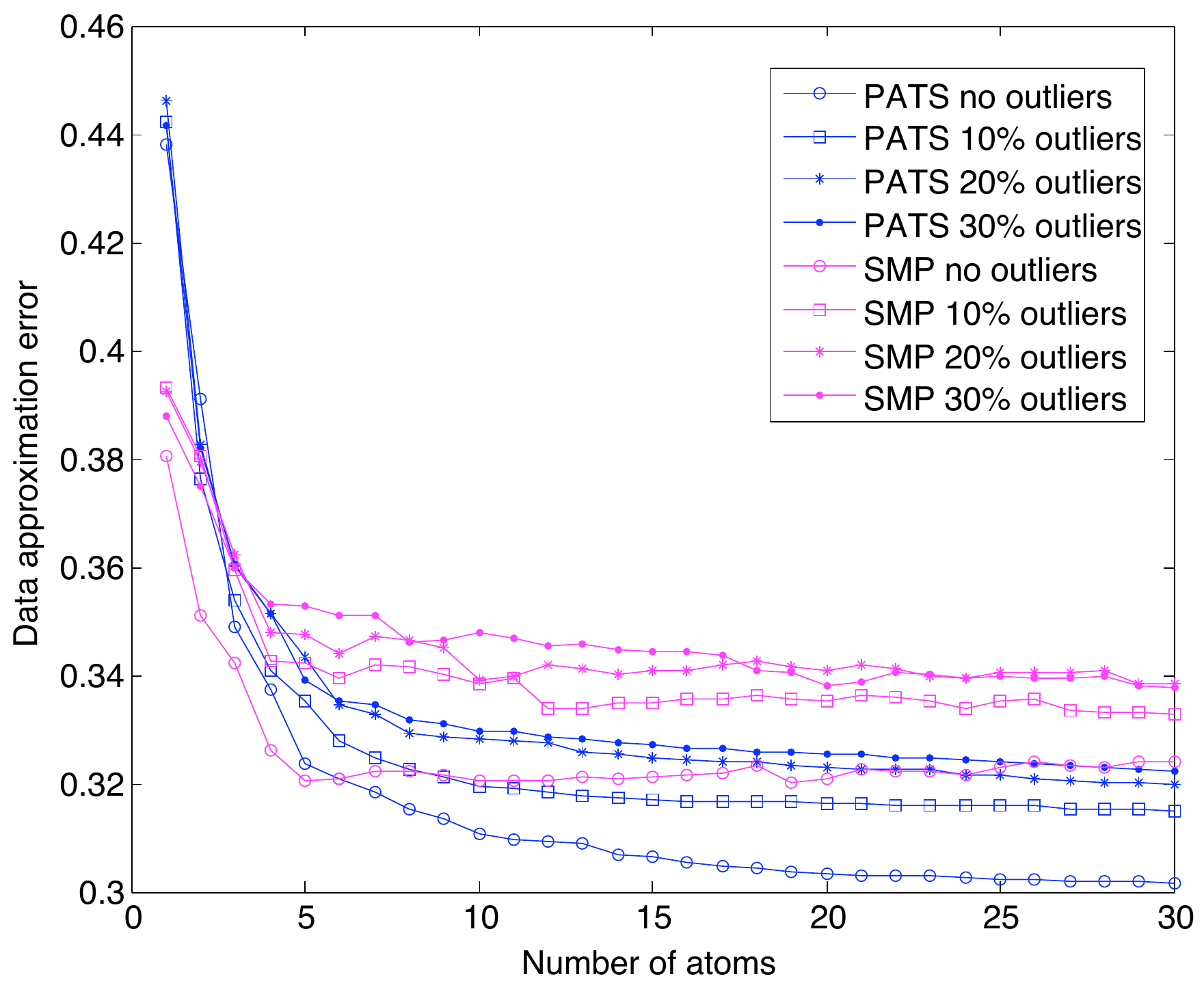}}
 \end{center}
 \caption{Manifold approximation results with occluded digit images with outliers}
 \label{fig:resultsDigitsTrans}
\end{figure}

}

\begin{figure}[]
\begin{center}
     \subfigure[Images from the Yale face data set]
       {\label{fig:databaseFacesScale}\includegraphics[height=1.6cm]{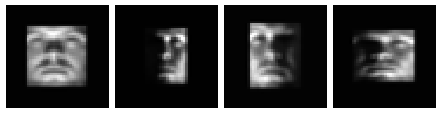}}
     \subfigure[Learned patterns. From left to right: Normal setting, without gradient descent, bad initialization.]
       {\label{fig:learnedFacesScale}\includegraphics[height=1.5cm]{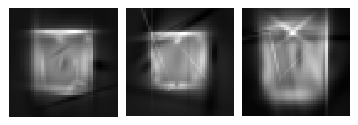}}      
      \subfigure[Approximation error]
       {\label{fig:errorsFaceScale}\includegraphics[width=8cm]{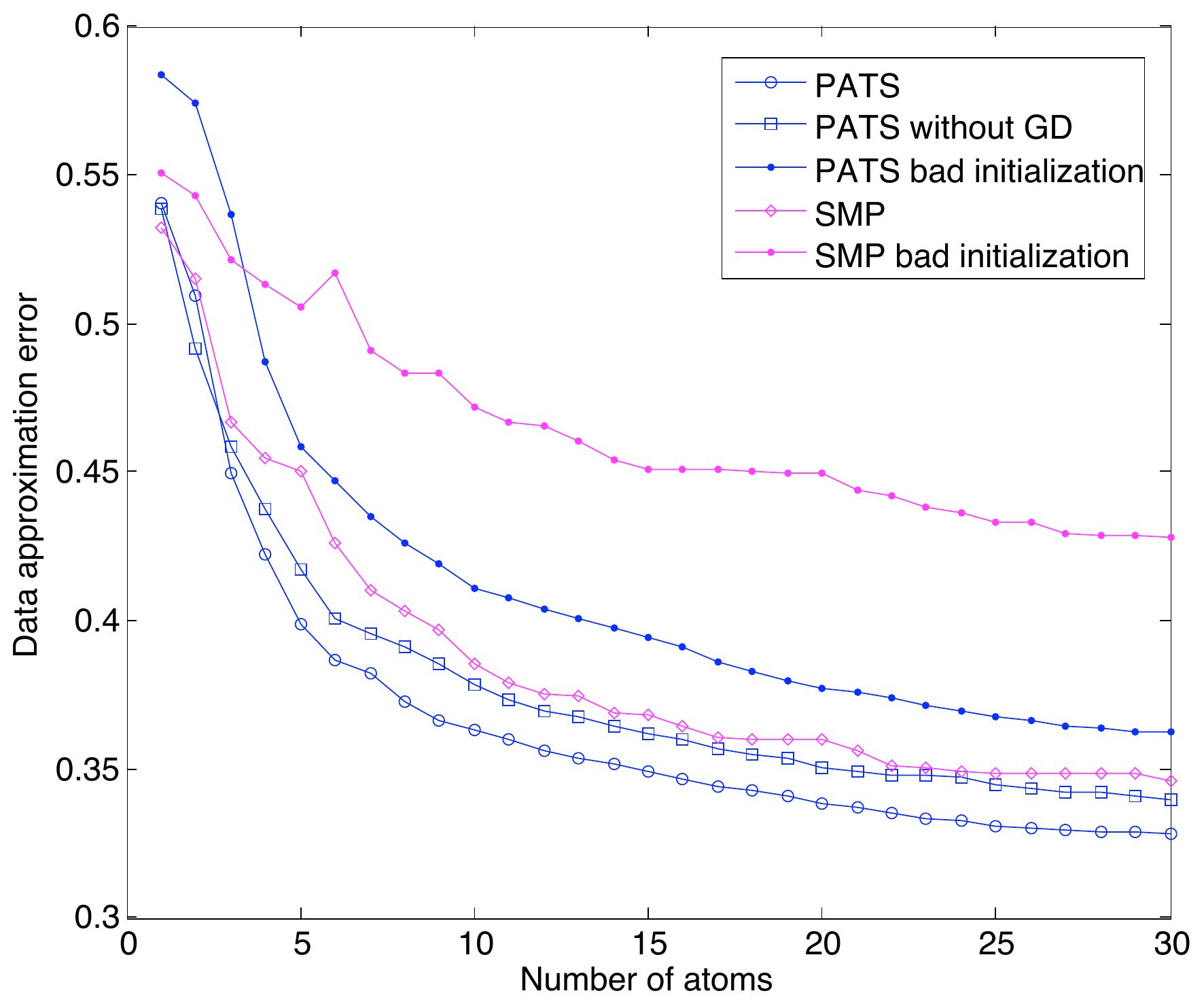}} 
 \end{center}
 \caption{Manifold approximation results with face images with varied illumination conditions}
 \label{fig:resultsFaceScale}
\end{figure}

\revis{Then, in a third experiment, we search the effect of some algorithm settings on the performance of PATS. We experiment on a data set from the Extended Yale Face Database B \cite{KCLee05} where face images are captured under varying illumination conditions. We create a data set of 90 images by applying geometric transformations consisting of anisotropic scaling to the images of a single subject, where only the parameters $s_x, s_y$ in (\ref{eq:manifoldModelSpecial}) are used. Some sample data set images are shown in Figure \ref{fig:databaseFacesScale}. We apply the PATS algorithm in three different settings. In the first setting, the algorithm is used in its normal mode; i.e., in line \ref{algPATS:lineInit} of the algorithm, parameters are initialized with respect to the data set image having the smallest distance to the centroid of all images. In the second setting, the initialization is done in the same way; however, line \ref{algPATS:lineGD} of the algorithm (gradient descent) is omitted. The third setting is the same as the first setting except that the algorithm is started with a bad initialization, where the alignment in line \ref{algPATS:lineInit} is done with respect to the data set image having the largest distance to the centroid. The patterns learned in all three settings are shown in Figure \ref{fig:learnedFacesScale}, and the approximation errors are plotted in Figure \ref{fig:errorsFaceScale}. The algorithm does not output clear facial features due to the variation of illumination. The gradient descent step is seen to bring a certain improvement in the performance. The results also show that the algorithm has a sensitivity to initialization. A significant change in the initial values of the transformation parameters causes the algorithm to compute a different solution. In order to provide a comparison, we also plot the results obtained with ``SMP on aligned patterns'' with default and bad alignments. The fact that the error difference between the two cases is much larger in SMP compared to PATS suggests that PATS can nevertheless compensate for a bad initialization of transformation parameters to some extent.}

\begin{figure}[t]
 \centering
  \includegraphics[scale=0.5]{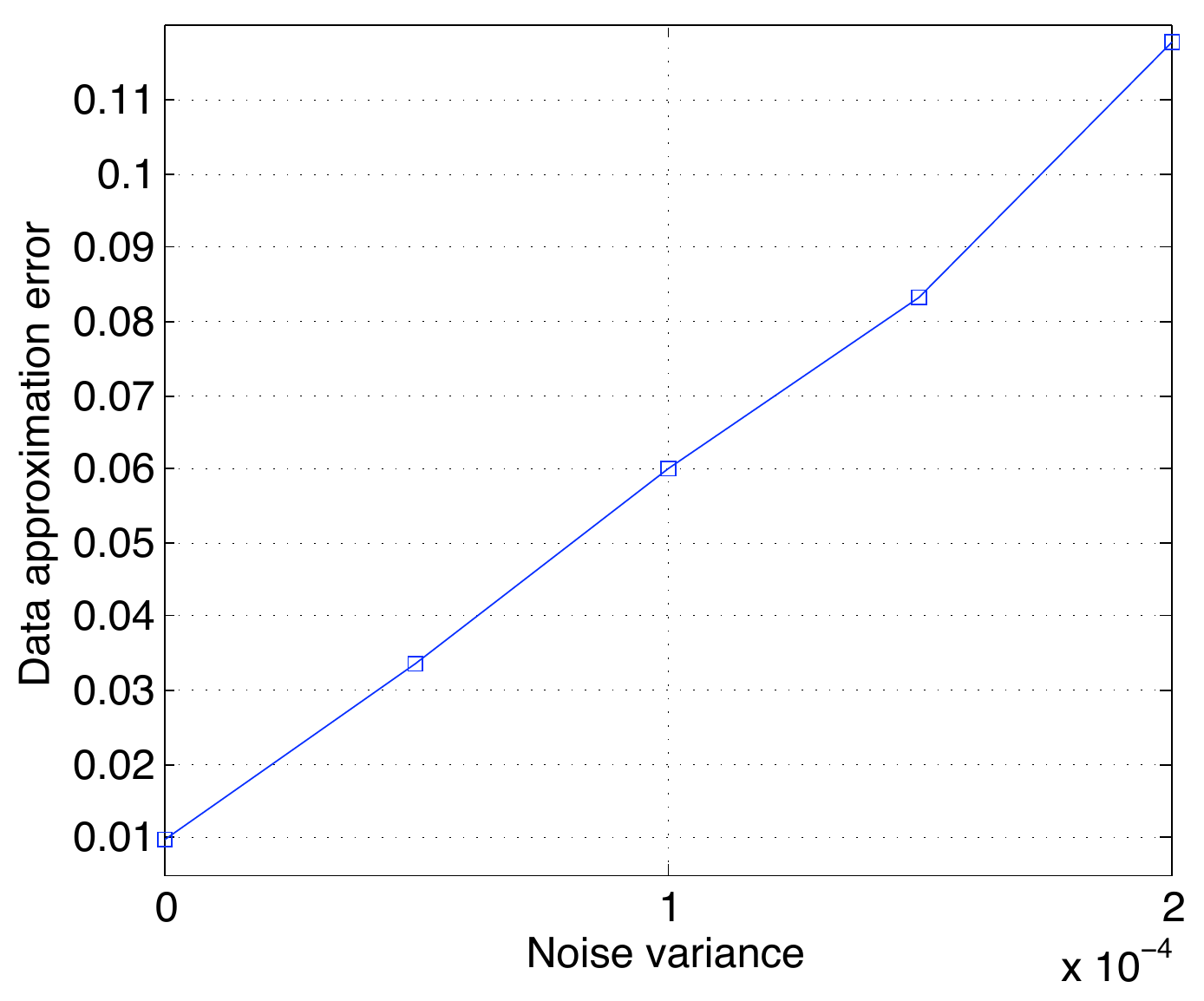}
  \caption{Dependence of the approximation error on the data noise. \revis{The largest noise variance $2\times 10^{-4}$ corresponds to an SNR of 9.054 dB.}}
  \label{fig:errorVsNoise}
\end{figure}

Finally, in a last experiment we examine the approximation accuracy of the learned manifold with respect to the noise level of the data set. We form a synthetic pattern $r$ that is composed of 10 randomly selected atoms from $\mathcal{D}$. Then, we generate a data set $\mathcal{U}$ of 50 images by applying to $r$ random geometric transformations of the form (\ref{eq:manifoldModelSpecial}). We derive several data sets from $\mathcal{U}$ by corrupting its images with additive Gaussian noise, where each data set has a different noise variance. Then, we run the PATS algorithm on each data set. In Figure \ref{fig:errorVsNoise}, the data approximation error is plotted with respect to the noise variance. The deviation between $\mathcal{U}$ and $\mathcal{M}(r)$ depends on the noise level, and the ideal approximation error is linearly proportional to the noise variance. Such a linear dependency can be observed in Figure \ref{fig:errorVsNoise}. However, one can note that the curve does not pass through the origin, which is due to the suboptimal greedy nature of the algorithm.

%
%
%
%

\section{Joint Computation of PTMs for Classification}
\label{sec:ClassBasedMB}

In this section we consider multiple image sets, where each set consists of geometrically transformed observations of a different signal class. We build on the scheme presented in Section \ref{ssec:appxPTMGeneric} and extend the PATS algorithm for joint manifold computation in classification applications.

\revis{We remark the following about the use of PTM models in transformation-invariant classification. Given a collection of PTMs representing different classes, for each manifold one can identify a subset of the whole image space that consists of points whose distances to that manifold are smaller than their distances to the other manifolds. We call this subset of the image space as the ``approximation region'' of the manifold (see \cite{vuralManifDisc}, Section II for a more detailed discussion). Note that it is not always possible to partition the whole image space into the approximation regions of a set of class-representative PTMs. For instance, one may come across degeneracies resulting from manifold intersections; or there may exist a full-dimensional subset of the image space that is equidistant to two manifolds. Yet, our PTM computing approach relies on the implicit assumption that in a transformation-invariant classification application, the training and test signals that belong to a certain class are in practice likely to be located close to the approximation region of a PTM.}

\subsection{Problem Formulation}
\label{ssec:ClassProblemForm}

Consider a collection of visual signals $\mathcal{U}=\bigcup_{m=1}^{M} \mathcal{U}^m \subset \mathbb{R}^n$ consisting of $M$ classes, where each subset $\mathcal{U}^m = \{ u_i^m \}_{i=1}^{N_m} $ consists of $N_m$ geometrically transformed observations of a visual signal of class $m$. We would like to represent each set $\mathcal{U}^m$ by a transformation manifold $ \mathcal{M}(p^m)$ that is generated by the geometric transformations of a representative pattern $p^m$. Let us denote 
\begin{equation}
\mathcal{M}^m=\mathcal{M}(p^m)=\{ U_{\lambda}(p^m), \lambda \in \Lambda \} \subset \mathbb{R}^n.
\label{eq:manModelSpecMult}
\end{equation}

We would like to build $ \{ \mathcal{M}^{m} \}$ such that they provide a good representation of the images in $\mathcal{U}$ and also permit to classify them accurately by manifold distance computation. Hence, in the construction of the manifolds, we formulate the objective function as a weighted combination of two terms $E_a$ and $E_c$, which respectively represent approximation and classification errors. The approximation error $E_a$ is given by the sum of the squared distances of images to the manifold of the same class

\begin{equation}
\label{eq:appErrorJoint}
E_a=\sum_{m=1}^M \sum_{i=1}^{N_m} \| e^m_i \|^2 =\sum_{m=1}^M \sum_{i=1}^{N_m}  d^2 ( u^m_i,  \mathcal{M}^m ).
\end{equation}

We assume that an image is assigned the class label of the manifold with smallest distance to it. We define a misclassification indicator function $I$ such that for $u_i^m \in \mathcal{U}^m $

\begin{equation}
I(u_i^m)=\bigg \{
\begin{array} {l}
0, \, \, \text{    if   } d(u_i^m,\mathcal{M}^m) < \min_{r \neq m} d\big(u_i^m, \mathcal{M}^r \big) \\ 
1, \, \, \text{    otherwise.}  \end{array}
\end{equation}
Then, the classification error $E_c$ is the total number of misclassified data points.

\begin{equation}
E_c= \sum_{m=1}^M \sum_{i=1}^{N_m} I (u_i^m)
\label{eq:orgClassErrorE}
\end{equation}
We would like to compute $\{ \mathcal{M}^m \}_{m=1}^M$ such that the weighted error 
\begin{equation}
E= E_a+ \alpha  \, E_c
\end{equation}
is minimized, where $\alpha >0$ is a coefficient adjusting the weight between the approximation and classification terms. We formulate a generic PTM learning problem as follows. \\

\begin{problem}
\label{prob:ClassGen}
Given image sets  $\{ \mathcal{U}^m \}$, compute patterns $\{ p^m \} \subset L^2(\mathbb{R}^2)$ and transformation parameters $\{ \lambda_i^m \} \subset \Lambda$, $m=1, \, \dots \,, M$ and $i=1, \, \dots \,, N_m$, by minimizing
\begin{eqnarray}
\begin{split}
E &= \sum_{m=1}^M \sum_{i=1}^{N_m}   \left(   \| u^m_i - U_{\lambda^m_i} ( p^m )  \|^2 + \alpha \, I (u_i^m) \right).
\end{split}
\end{eqnarray}
\end{problem}

Our solution is based on constructing each $p^m$ using atoms from the dictionary manifold $\mathcal{D}$ defined in (\ref{eq:dictManifold}). We reformulate Problem \ref{prob:ClassGen} under these assumptions.\\

\begin{problem}
\label{prob:ClassDict}
Given image sets  $\{ \mathcal{U}^m \}$, a mother function $\phi$ and sparsity constraints $\{K_m\}$; compute a set of atom parameters $\{ \gamma^m_j \} \subset \Gamma$, coefficients $\{c^m_j\} \subset \mathbb{R}$, and transformation parameters $\{ \lambda_i^m \} \subset \Lambda$ for $m=1, \, \dots \,, M$, $j=1, \, \dots \,, K_m$ and $i=1, \, \dots \,, N_m$, by minimizing
\begin{eqnarray}
\begin{split}
E &= \sum_{m=1}^M \sum_{i=1}^{N_m}   (   \| u^m_i - U_{\lambda^m_i} \big( \sum_{j=1}^{K_m} c^m_j  \phi_{\gamma^m_j}  \big)  \|^2 + \alpha \, I (u_i^m) ).
\end{split}
\label{eq:probFrmAtClas}
\end{eqnarray}
\end{problem}

\subsection{Classification-Driven PTM Learning}
\label{ssec:classPTM}

Problem \ref{prob:ClassDict} is similar to Problem \ref{prob:AppxDict}, except that it also involves a classification error term that has a quite complex dependence on the optimization variables. Therefore, it is hard to solve optimally. We present a constructive solution based on building $\{p^m\}$ iteratively with joint atom selection.

We begin with a tentative assignment of parameter vectors. In (\ref{eq:probFrmAtClas}) each vector $\lambda_i^m$ corresponds to the projection of $u_i^m$ on $\mathcal{M}^m$. We assign $\{ \lambda_i^m \}$ by picking a reference pattern $\Psi^m$ for each class and then projecting each $\mathcal{U}^m$ onto $\mathcal{M}(\Psi^m)$. We also compute the cross-projection vectors $\{ \lambda^{m , r}_i \}$, where
\begin{equation*}
 \lambda^{m , r}_i = \arg \min_{\lambda \in \Lambda} \|  u_i^m - U_{\lambda}(p^r)  \|
\end{equation*}
corresponds to the projection of $u^m_i$ onto $\mathcal{M}^r$.

Then, we construct $\{p^m\}$ by gradually adding new atoms to each $ p^m $.  In the $j$-th iteration of the algorithm, we would like to optimize the parameters ${\gamma}^m_j$ and coefficients $c^m_j$ of the new atoms such that the weighted error $E$ is minimized. Now we consider the $j$-th iteration and denote $\gamma^m=\gamma^m_j$, $c^m=c^m_j$. Then $\gamma=[ \gamma^1 \,  \gamma^2  \, \dots \,  \gamma^M ]$ and $c=[ c^1 \, c^2 \, \dots \,  c^M ]$ are the optimization variables of the $j$-th iteration. We consider $E$ as a function of $\gamma$ and $c$ similarly to Section \ref{sec:AppxBasedMB} and propose to minimize $E$ through a two-stage optimization. We first obtain an approximation $\tilde E$ of $E$, which is in a DC form.  We minimize $\tilde E $ using the cutting plane algorithm and estimate a coarse solution, which is used as an initial solution in the second stage. Then in the second stage, we define a refined approximation $\hat {E}$ of $E$ based on the tangent distances of images to the manifolds and minimize it with a gradient-descent algorithm. 

The minimization of $\tilde{E}$ and $\hat E$ determines a solution for $\gamma$ and $c$. We update the pattern $p^m$ of each class by adding it the selected atom with parameters $\gamma^m$ and coefficient $c^m$ (\revistwo{in practice, we add an atom only if its coefficient is significant enough}). Then, we recompute the transformation parameters $\{ \lambda_i^{m}\}$ and $\{ \lambda_i^{m ,r}\}$ by projecting the images onto the new manifolds. \revis{We have observed that selecting the atoms by minimizing a combination of approximation and classification terms instead of only a classification term gives better results, especially for robustness to data noise. Still, we would like to make sure that the selected atoms improve the classification performance at the end of an iteration. Therefore, the decision of accepting the updates on the manifolds is taken according to the classification error $E_c$ in (\ref{eq:orgClassErrorE}). If $E_c$ is not reduced we reject the updates and pass to the next iteration.}\footnote{\revis{In the course of the algorithm, parameters $\beta$ and $\alpha$ are adapted such that the emphasis is shifted from approximation capabilities in early phases to classification capabilities in later phases. This is explained in more detail in Section \ref{ssec:classPTMImpl}. For this reason, even if the classification error does not decrease in one iteration, it may do in the next one.}} We continue the iterations until the classification error $E_c$ converges. The \revis{termination} of the algorithm is guaranteed by constraining $E_c$ to be non-increasing during the iterations, which in return stabilizes the objective function $E$. We call this method Joint Parameterized Atom Selection (JPATS) and summarize it in Algorithm \ref{alg:jpats}.

Let us come to the detailed description of the approximations of $E$ in the two-stage optimization. Firstly, let $\{ p^m_{j-1} \}$ and $\{ \mathcal{M}^m_{j-1} \}$ denote the patterns and the corresponding transformation manifolds computed after $j-1$ iterations. For simplicity of notation, we will use the convention $\mathcal{M}^m = \mathcal{M}^m_{j}$ and $p^m = p^m_j$ throughout the derivations of $\tilde E$ and $\hat E$.

In the first step, we obtain $\tilde E$ in the form $\tilde E = \tilde {E}_a + \alpha \, \tilde {E}_c$, where $\tilde {E}_a$ and $\tilde {E}_c$ are respectively the approximations of  $E_a$ and $E_c$. The first term $\tilde {E}_a$ is simply given by the generalization of the approximation error in (\ref{eq:tildeE}) to the multiple manifold case. 
\begin{equation}
\label{eq:tildeEClass}
\tilde{E}_a = \sum_{m=1}^M   \sum_{i=1}^{N_m}  \| \tilde{e}^m_i \|^2 
	    =  \sum_{m=1}^M   \sum_{i=1}^{N_m}  \| v^m_i -  c^m \, U_{\lambda^m_i} ( \phi_{\gamma^m} )  \|^2 
\end{equation}
where the parameters $\lambda^m_i$ are the ones computed at the end of iteration $(j-1)$, and $v^m_i = u^m_i - U_{\lambda^m_i}(p^m_{j-1} )$.

Then, we derive $\tilde {E}_c$ in the following way. Notice that the classification error $E_c$ in (\ref{eq:orgClassErrorE}) is a discontinuous function of $\gamma$ and $c$ due to the discontinuity of the misclassification indicator function $I$. Let 
$r(u_i^m)$ denote the index of the manifold with smallest distance to an image $u_i^m$ among the manifolds of all classes except its own class $m$; i.e.,
\begin{equation*}
r(u_i^m)= \arg \min_{r\neq m} d(u_i^m, \mathcal{M}^r).
\end{equation*}
It is clear that $r(u_i^m)$ can take different values throughout the iterations. However, for simplicity, in the $j$-th iteration we fix the indices $r(u_i^m)$ to their values attained at the end of iteration $(j-1)$ and denote them by the constants $r_i^m$. Then we can define the function

\begin{equation*}
f(u_i^m)=d^2(u_i^m,\mathcal{M}^m) - d^2(u_i^m,\mathcal{M}^{r_i^m} )
\end{equation*}
such that $I(u^m_i)$ corresponds to the unit step function of $f(u_i^m)$; i.e., $I(u^m_i)=u(f(u_i^m))$. Thus, if we replace the unit step function with the sigmoid function $S(x)=(1+e^{-\beta x})^{-1}$, which is a common analytical approximation of the unit step, we obtain the approximation 
$
S\big(f(u_i^m)\big)=(1+e^{- \beta f(u_i^m)} )^{-1}
$
of $I(u_i^m)$.
As the value of the positive scalar $\beta$ tends to infinity, the sigmoid function approaches the unit step function. A continuous approximation of $E_c$ is thus given by 
\begin{equation}
\sum_{m=1}^M \sum_{i=1}^{N_m} S\big(f(u_i^m)\big).
\label{eq:cerror_sumS}
\end{equation}
Now, in order to minimize the function in (\ref{eq:cerror_sumS}) we do the following. We first compute
\[ f_0(u_i^m)=d^2(u_i^m,\mathcal{M}_{j-1}^m) - d^2(u_i^m,\mathcal{M}_{j-1}^{r_i^m}) \]
for each image $u_i^m$. Then, applying a first-order expansion of $S$ around each $f_0(u_i^m)$, we obtain the following approximation of the error term in (\ref{eq:cerror_sumS}).
\begin{eqnarray}
\label{eq:cerror_firstOrdsumS}
\sum_{m=1}^M \sum_{i=1}^{N_m}  \bigg( S\big(f_0(u_i^m)\big) 
                           + \frac{d S}{d f} \bigg|_{f=f_0(u_i^m)}   \big( f(u_i^m) - f_0(u_i^m) \big) \bigg)
\end{eqnarray}
Since $f_0(u_i^m)$ and $S(f_0(u_i^m))$ are constants, the minimization of the expression in (\ref{eq:cerror_firstOrdsumS}) becomes equivalent to the minimization of
\begin{eqnarray}
\sum_{m=1}^M \sum_{i=1}^{N_m}   \frac{d S}{d f} \bigg|_{f=f_0(u_i^m)}   f(u_i^m)   = \sum_{m=1}^M \sum_{i=1}^{N_m}  \eta_i^m f(u_i^m)
\label{eq:cerror_nus1}
\end{eqnarray}
where
\begin{equation*}
\eta_i^{m} = \frac{d S}{d f} \bigg|_{f=f_0(u_i^m)} =\frac{\beta \, e^{-\beta f}}{(1+e^{-\beta f })^2 }\bigg|_{f=f_0(u_i^m)}.
\end{equation*}

Let us rearrange (\ref{eq:cerror_nus1}) in a more convenient form. For each class index $m$, let $R^m=\{ (i,k): r_i^k=m \}$ consist of the pairs of data and class indices of images that do not belong to class $m$ but have $\mathcal{M}^m$ as their closest manifold among all manifolds except the one of their own class. Then (\ref{eq:cerror_nus1}) can be rewritten as 
\begin{equation}
\begin{split}
  \sum_{m=1}^M   \sum_{i=1}^{N_m}  \eta_i^{m } \, d^2(u_i^m,\mathcal{M}^m)  
     -  \sum_{m=1}^M \sum_{(i,k)\in R^m}  \eta_i^{k}  \, d^2(u_i^k,\mathcal{M}^{m}) .
\label{eq:cerror_nus2}
\end{split}
\end{equation}

As it is not easy to compute the distance terms $d^2(u_i^k,\mathcal{M}^m) $ directly, we proceed with the approximation $d^2(u_i^k,\mathcal{M}^m) \approx  \| u_i^k  -  U_{\lambda^{k,m}_i} (  p^m_{j-1}  +   c^m \,\phi_{\gamma^m} )  \|^2$, where the value of ${\lambda^{k , m}_i}$ is the one computed in iteration $(j-1)$. We finally get $\tilde{E}_c$ from (\ref{eq:cerror_nus2}) with this approximation.
\begin{equation}
\label{eq:hatEpsilon1}
\begin{split}
\tilde E_c &=\sum_{m=1}^M   \sum_{i=1}^{N_m}  \eta_i^{m }  \,  \| v_i^m  -  c^m  \, U_{\lambda^m_i} ( \phi_{\gamma^m} )  \|^2 \\
             & - \sum_{m=1}^M  \sum_{(i,k)\in R^m}  \eta_i^{k} \,  \| v^{k , m}_i- c^m \, U_{\lambda^{k , m}_i} ( \phi_{\gamma^m} ) \|^2
\end{split}
\end{equation}
where $v^{k , m}_i = u^k_i - U_{\lambda^{k , m}_i}(p^m_{j-1} )$. Now, from (\ref{eq:tildeEClass}) and (\ref{eq:hatEpsilon1}) we can define
\begin{equation}
\label{eq:tildeGform}
\tilde E = \tilde{E}_a+ \alpha \, \tilde{E}_c.\\
\end{equation}

\begin{proposition}
\label{prop:DCclassError}
$\tilde E$ is a DC function of $\gamma$ and $c$. Moreover, if a DC decomposition for the components of the transformed atom $U_{\lambda} (  \phi_{\gamma} )$ is known, a DC decomposition of $\tilde{E}$ is computable.\\
\end{proposition}

The proof of Proposition \ref{prop:DCclassError} is given in Appendix D.\\

Now let us describe the term $\hat E$ that is used in the second stage of the optimization of $E$.  We derive $\hat E$ by replacing the manifold distances by tangent distances; i.e., we use the approximation $d^2(u_i^k,\mathcal{M}^m) \approx d^2 (u_i^k, \mathcal{S}^{k}_i(p^m))$, where $\mathcal{S}^{k}_i(p^m)$ is the first-order approximation of $\mathcal{M}^m$ around the point $U_{\lambda^{k , m}_i} (p^m)$. The tangent distance is derived in Appendix B. Let $w^m_i = u^m_i - U_{\lambda^m_i}(p^m)$ and  $w^{k , m}_i = u^k_i - U_{\lambda^{k , m}_i}(p^m )$. Then the function $E_a$ in (\ref{eq:appErrorJoint}) is approximated by
\begin{equation}
\label{eq:hatEa}
\hat{E}_a=  \sum_{m=1}^M   \sum_{i=1}^{N_m}   \| w^m_i  - {T_{i}^m}  \left( ({T_{i}^m})^{\mbox{\scriptsize{T}}} \, {T_{i}^m} \right)^{-1} \, ({T_{i}^m})^{\mbox{\scriptsize{T}}} w^m_i   \|^2.
\end{equation}
Similarly, the classification error function in (\ref{eq:cerror_nus2}) is approximated by
\begin{eqnarray}
\begin{split}
\hat{E}_c &= \sum_{m=1}^M   \sum_{i=1}^{N_m}  \eta_i^{m }    
                                \| w^m_i  - {T_{i}^m}  \left( ({T_{i}^m})^{\mbox{\scriptsize{T}}} \, {T_{i}^m} \right)^{-1} \, ({T_{i}^m})^{\mbox{\scriptsize{T}}} w^m_i   \|^2  \\
                            & - \sum_{m=1}^M  \sum_{(i,k)\in R^m}  \bigg( \eta_i^{k}  \\
                            & \cdot  \| w^{k , m}_i  - {T_{i}^{k , m}}  \left( ({T_{i}^{k , m}})^{\mbox{\scriptsize{T}}} \, {T_{i}^{k , m}} \right)^{-1} \, ({T_{i}^{k , m}})^{\mbox{\scriptsize{T}}} w^{k , m}_i   \|^2 \bigg).
\label{eq:hatEpsilon2}
\end{split}
\end{eqnarray}
Here $ {T_{i}^m}$ and $ {T_{i}^{k , m}}$ denote the $n \times d$ matrices whose columns are the tangent vectors to the manifold $\mathcal{M}^m$ at respectively the points $U_{\lambda^m_i}(p^m) $ and  $U_{\lambda^{k , m}_i}(p^m) $. From (\ref{eq:hatEa}) and (\ref{eq:hatEpsilon2}) we can finally define
\begin{equation}
\label{eq:hatGform}
\hat {E}= \hat {E}_a+ \alpha \, \hat {E}_c.
\end{equation}

\revis{Let us briefly discuss the effect of the approximations made on the original cost function $E$. The accuracy  of approximating the unit step function with a sigmoid in (\ref{eq:cerror_sumS}) can be adjusted by changing the slope of the sigmoid (see also the note in Section  \ref{ssec:classPTMImpl}). Then, in order for the linear approximation of the sigmoid in (\ref{eq:cerror_firstOrdsumS}) to be valid, the values of $f(u_i^m)$ must be sufficiently close to their base values $f_0(u_i^m)$. The effect of this linearization can be alleviated by updating the base values $f_0(u_i^m)$ several times in an iteration. The rest of the approximations are similar to those discussed in Section \ref{ssec:appxPTMGeneric}.  }

\begin{algorithm}[h]
\caption{Joint Parameterized Atom Selection (JPATS)}

\begin{algorithmic}[1]

\STATE
\textbf{Input:} \\
$\mathcal{U}=\bigcup_{m=1}^{M} \mathcal{U}^m  $: Set of observations for $M$ signal classes\\

\STATE
\textbf{Initialization:}

\STATE
Determine tentative parameter vectors $\{ \lambda_i^{m , r}  \}$ by projecting $\{ u_i^m \}$ on the transformation manifolds $\{\mathcal{M}(\Psi^m) \}$ of reference patterns $\{ \Psi^m \}$.

\STATE
$p_0^m=0$ for $m=1, \, \dots \,, M$.

\STATE
$j=0$.

\STATE
Initialize the sigmoid parameter $\beta$ and the weight parameter $\alpha$.

\REPEAT

\STATE
$j=j+1$.

\STATE
\label{state:atomSelectJPATS}
Optimize the joint atom parameters $\gamma=[ \gamma^1 \,  \gamma^2  \, \dots \,  \gamma^M ]$ and coefficients $c=[ c^1 \, c^2 \, \dots \,  c^M ]$ with DC programming such that the error $\tilde{E}$ in (\ref{eq:tildeGform}) is minimized.

\STATE
\label{state:gradientDesJPATS}
Further optimize ${\gamma}$ and $c$ with gradient descent such that the refined error $\hat E$ in (\ref{eq:hatGform}) is minimized.

\STATE
Update $p^m_j=p^m_{j-1}+c^m \,  \phi_{\gamma^m}$ for $m=1, \, \dots \,, M$ if $c^m$ is significant.

\STATE
Update the parameter vectors $\{ \lambda_i^{m , r} \}$.

\STATE
Update $\beta$ and $\alpha$.

\STATE
\label{}
Check if the new manifolds reduce the classification error $E_c$. If not, reject the updates on $p^m$ and $\{ \lambda_i^{m , r} \}$, and go back to \ref{state:atomSelectJPATS}.

\UNTIL the classification error $E_c$ converges

\STATE
\textbf{Output}:\\
 $\{p^m\}  = \{p^m_j\}$: A set of patterns whose transformation manifolds $\{ \mathcal{M}^m \}$ represent the data classes $\mathcal{U}^m$\\

\end{algorithmic}
\label{alg:jpats}
\end{algorithm}

\subsection{Implementation Details}
\label{ssec:classPTMImpl}

We now discuss some points related to the implementation of JPATS. We first explain the choice of the parameter $\beta$ in Algorithm \ref{alg:jpats}. Notice that the function $S\big(f(u_i^m)\big)$ can also be interpreted as the probability of misclassifying $u_i^m$ upon updating the manifolds at the end of the iteration. When  $u_i^m$ gets closer to its true manifold $\mathcal{M}^m$, $f(u_i^m)$ decreases and $S\big(f(u_i^m)\big)$ decays to $0$. Similarly, when $u_i^m$ gets away from $\mathcal{M}^m$, $S\big(f(u_i^m)\big)$ approaches 1. The probabilistic interpretation of the function $S\big(f(u_i^m)\big)$ stems naturally from its shape. Consequently, the approximate error in (\ref{eq:cerror_sumS}) corresponds to the sum of the probabilities of misclassifying the input images. Based on this interpretation, we propose to update $\beta$ according to the statistics drawn from the data. For each $u_i^m$, we examine the value of $f(u_i^m)$ at the beginning the iteration and the value of $I(u_i^m)$ at the end of the iteration. Then we pick $\beta$ such that the shape of the sigmoid matches the $I(u_i^m)$ vs. $f(u_i^m) $ plot. Such an adaptive choice of $\beta$ also provides the following flexibility. In early phases of the process where the total misclassification rate is relatively high, $\beta$ usually has small values, which yields slowly changing sigmoids. Therefore, a relatively large portion of the input images have an effect on the choice of the new atoms. However, in later phases, as the total misclassification rate decreases, $\beta$ usually takes larger values resulting in sharper sigmoids, which gives misclassified images more weight in atom selection.

Then, we comment on the choice of the weight parameter $\alpha$. In principle, $\alpha$ can be set to have any nonnegative value. Setting $\alpha=0$ corresponds to a purely approximation-based procedure that computes the manifolds individually with PATS, whereas a large $\alpha$ yields a learning algorithm that is rather driven by classification objectives. However, we have observed that a good choice in practice consists of selecting a small value for $\alpha$ at the beginning and increasing it gradually.\footnote{\revis{In our setup, we control the $\alpha$ parameter by using a shifted and scaled sigmoid function. The initial and final values of the sigmoid are around 0.5 and 10; and its center is typically attained at iterations 5-7 of Algorithm \ref{alg:jpats}.}} This guides the algorithm to first capture the main characteristics of input signals, and then encourage the selection of features that ensure better class-separability. 

Finally, we have made the following simplification in the implementation of the DC programming block. The number of optimization variables is $(s+1)M$ in our problem, where $s$ is the dimension of $\mathcal{D}$ and $M$ is the number of classes. Although the cutting plane algorithm works well for low-dimensional solution spaces, it becomes computationally very costly in high dimensions. Therefore, in the implementation of JPATS we partition the variables into subsets and optimize the subsets one by one.  Although there is no guarantee of finding the globally optimal solution in this case, we have experimentally observed that one can still obtain reasonably good results regarding the complexity-accuracy tradeoff. In order to handle high-dimensional solution spaces, one can alternatively replace the cutting plane algorithm with another DC solver such as DCA \cite{Tao97} or CCCP \cite{Yuille02theconcave}. These methods reduce the original DC program to the iterative solution of a pair of dual convex programs, which improves the computational complexity significantly at the expense of losing global optimality guarantees. Another issue affecting the efficiency of the DC programming block is the size of the solution space. We have seen that  it is useful to add a preliminary block that locates a good search region before the DC block. This can be achieved using a coarse grid in the solution space or a global search method such as the genetic algorithm or particle swarm optimization. \revis{Note that one may also minimize the objective function by using only a global search method. However, in experiments we have seen that the final value of the objective function is the smallest when both global search and DC optimization are employed.}

\subsection{Experimental Results}
\label{ssec:expClassPTM}

We now evaluate the performance of JPATS with experiments on transformation-invariant classification. We test the algorithm on two data sets consisting of handwritten digits \cite{lecun98} and microbiological images \cite{NHMpage}. In the digits experiment, we use the transformation manifold model in (\ref{eq:manifoldModelSpecial}). In the microbiological images experiment, we use the model
\begin{equation}
\mathcal{M}(p)=\{ U_{\lambda}(p): \lambda = (\theta, t_x, t_y, s) \in \Lambda \} \subset \mathbb{R}^n,
\label{eq:manifoldModelBact}
\end{equation}
where $s$ denotes an isotropic scale change. In both experiments, we use the dictionary model in (\ref{eq:dictManifoldExp}) and the Gaussian mother function in (\ref{eq:gaussMotherfunc}).

The first experiment is conducted on the images of the ``2,3,5,8,9'' digits, which lead to a relatively higher misclassification rate than the rest of the digits. The data sets are generated by randomly selecting 200 training and 200 test images for each digit and applying random geometric transformations consisting of rotation, anisotropic scaling and translation. The images of each digit are considered as the observations of a different signal class.

The second experiment is done on some sequences from the microbiology video collection of the Natural History Museum \cite{NHMpage}, which contains short video clips of living protists. We run the experiment on 6 different species (Discocephalus sp., Epiclintes ambiguus, Oxytricha sp., Scyphidia sp., Stentor roeseli, Stylonychia sp.), and we use three sample videos for each one. Each species is considered as a different class. The manifold in (\ref{eq:manifoldModelBact}) provides a suitable model, as the rotation and translations describe well the movements of the protists, and the isotropic scaling compensates for zoom changes. However, there is still some deviation from the manifold, as a result of noise, small nonrigid protist articulations and occasional recording of different individuals in different videos.  For each species, we experiment on a subset of frames from all three sequences. We preprocess the frames by conversion to greyscale, smoothing and thresholding. Then, for each class, we randomly select 70 training and 35 test images.

\begin{figure}[]
\begin{center}
     \subfigure[Sample images from data set]
       {\label{fig:databaseDigitClass}\includegraphics[width=7cm]{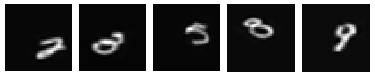}}
     \subfigure[Patterns built with JPATS]
       {\label{fig:classDigitsJPATS}\includegraphics[width=7cm]{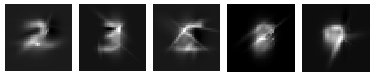}}
      \subfigure[Patterns built with PATS]
       {\label{fig:classDigitsPATS}\includegraphics[width=7cm]{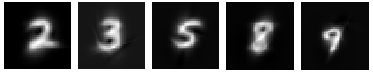}}
        \subfigure[Patterns built with SMP on aligned patterns]
       {\label{fig:classDigitsSMP}\includegraphics[width=7cm]{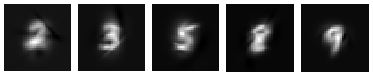}}
        \subfigure[Patterns built with SAS on aligned patterns]
       {\label{fig:classDigitsSAS}\includegraphics[width=7cm]{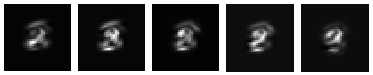}}
      \subfigure[Classification errors of test images]
       {\label{fig:errorDigitClass}\includegraphics[width=7cm]{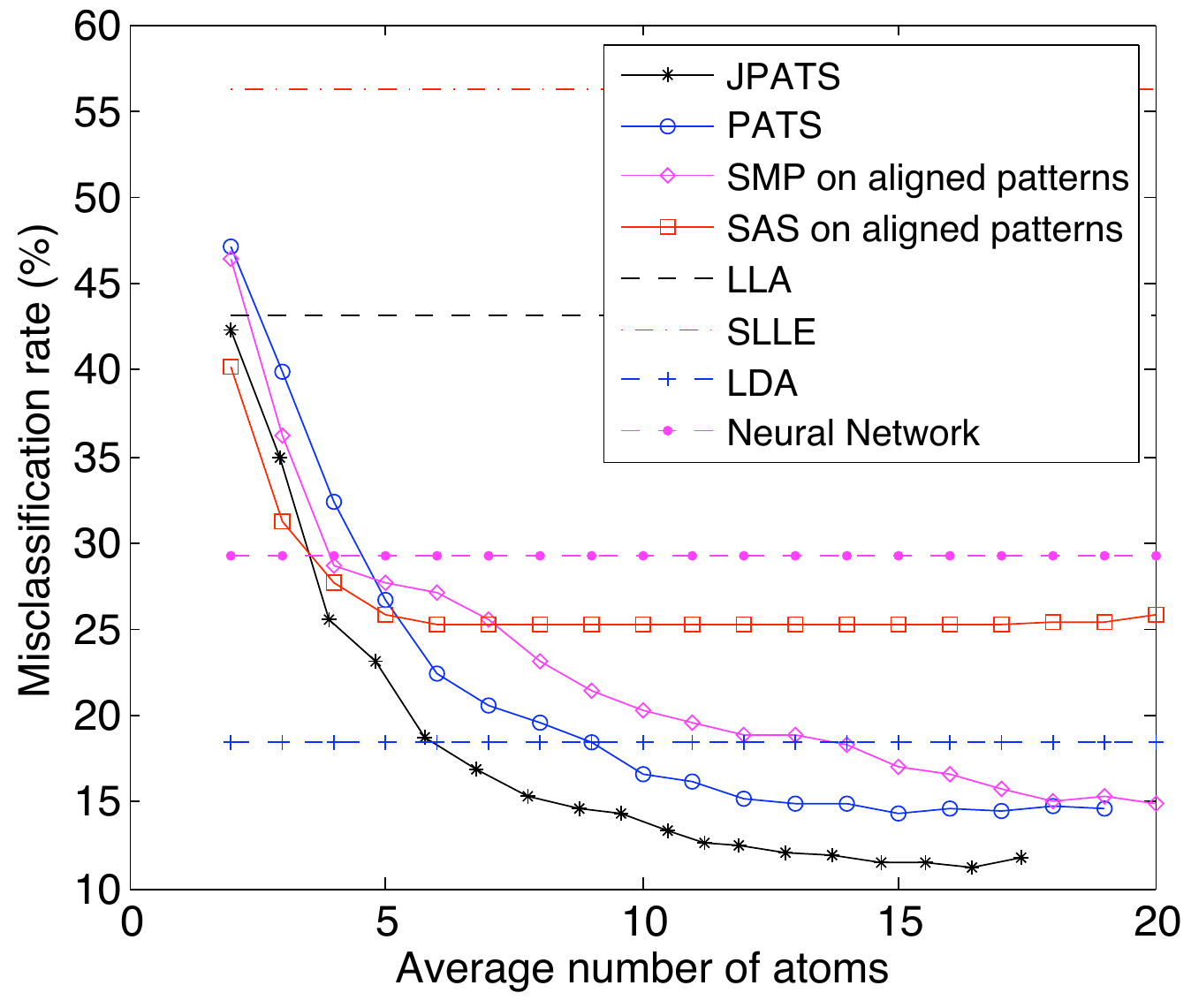}} 
 \end{center}
 \caption{Performance of the classification-driven learning algorithms on the handwritten digits data set}
 \label{fig:resultsDigitClass}
\end{figure}

\begin{figure}[]
\begin{center}
     \subfigure[Sample images from data set]
       {\label{fig:databaseBactClass}\includegraphics[width=8cm]{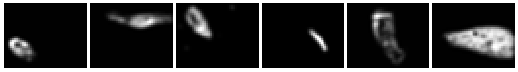}}
     \subfigure[Patterns built with JPATS]
       {\label{fig:classBactJPATS}\includegraphics[width=8cm]{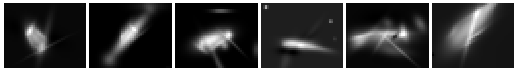}}
      \subfigure[Patterns built with PATS]
       {\label{fig:classBactPATS}\includegraphics[width=8cm]{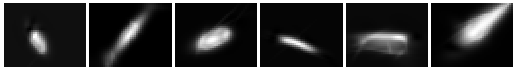}}
        \subfigure[Patterns built with SMP on aligned patterns]
       {\label{fig:classBactSMP}\includegraphics[width=8cm]{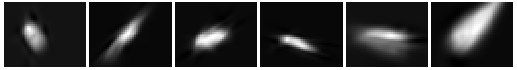}}
        \subfigure[Patterns built with SAS on aligned patterns]
       {\label{fig:classBactSAS}\includegraphics[width=8cm]{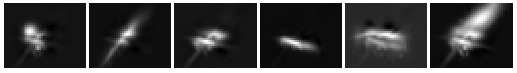}}
      \subfigure[Classification errors of test images]
       {\label{fig:errorBactClass}\includegraphics[width=7cm]{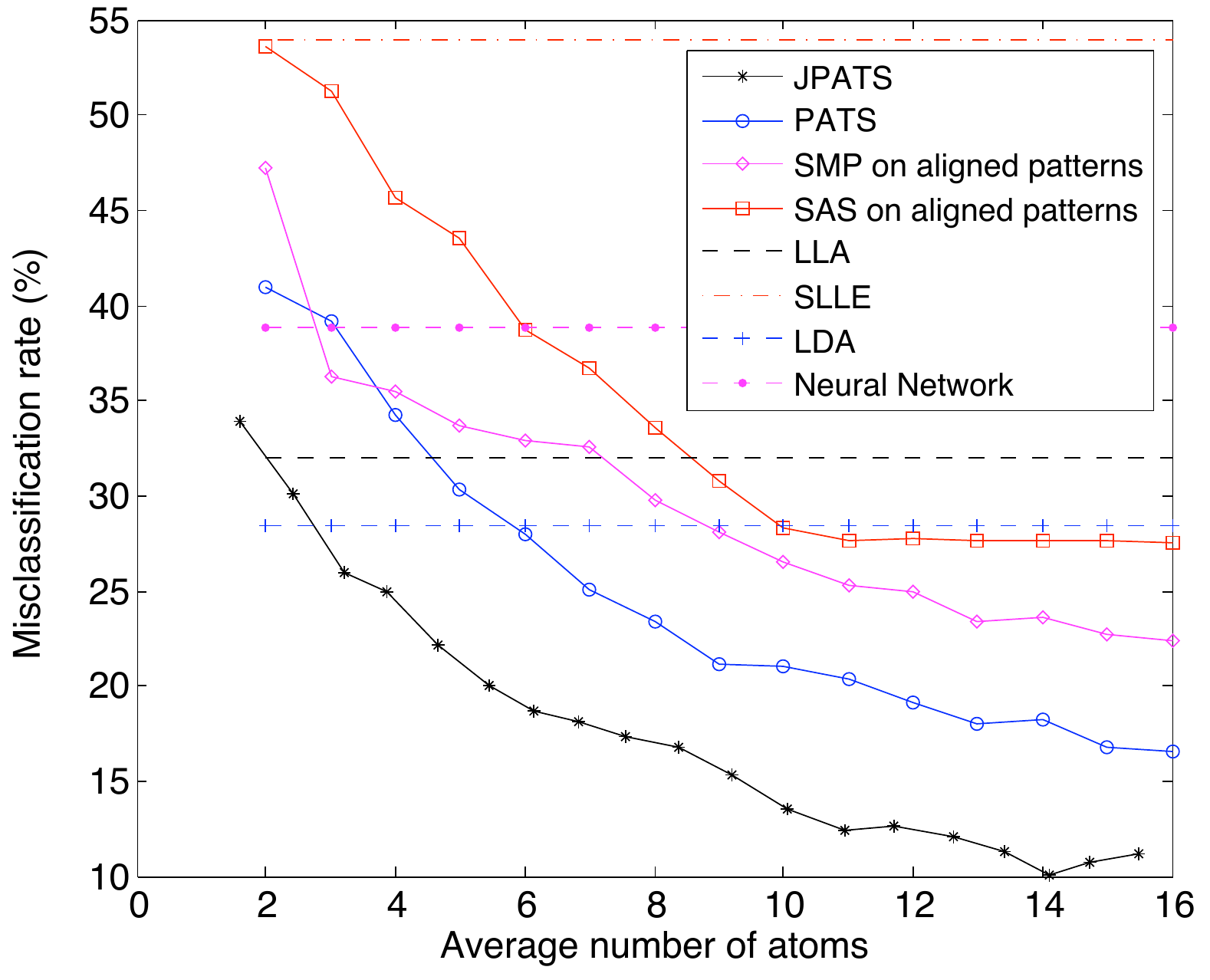}} 
 \end{center}
 \caption{Performance of the classification-driven learning algorithms on the microbiological images data set}
 \label{fig:resultsBactClass}
\end{figure}

In the experiments we compare the methods listed below. In the first four methods, we apply the algorithms on the training images in order to build PTMs. Then we compute the misclassification rate of the test images. The class label of a test image is estimated by identifying the smallest distance between the image and the computed manifolds. The  algorithms work as follows.

\begin{itemize}

\item JPATS: We jointly build PTMs for all classes with the proposed method.

\item PATS: We compute individual PTMs for each class with PATS.

\item SMP on aligned patterns: We compute individual PTMs for each class as explained in Section \ref{ssec:appxExperiments}.

\item SAS on aligned patterns: We use the untransformed/aligned images of all classes and select a set of Gaussian atoms with SAS  \cite{EPFLARTICLE91054}. \revistwo{We set the weight factor to $\lambda=2$ in \cite{EPFLARTICLE91054}.} Then, for each class we build a PTM by forming a pattern, where the selected atoms are weighted with their average coefficients.

\item LLA: We compute a locally linear approximation using the training images of each class. A test image is classified by identifying its $(d+1)$-nearest neighbors among the training images of each class, computing its distance to the plane passing through the nearest neighbors, and comparing its distances to the planes of different classes.

\item SLLE: We compute a low-dimensional embedding of the training images with the Supervised Locally Linear Embedding algorithm \cite{deridder:supervised}
and assign the class labels of the test images via nearest-neighbor classification in the embedded domain.

\revis{
\item LDA: Linear Discriminant Analysis on aligned data. The better one of linear and quadratic kernels is picked in each experiment.

\item Neural Network: A feed-forward backpropagation network for pattern recognition is used on aligned data.
}
\end{itemize}

The results are presented in Figures \ref{fig:resultsDigitClass} and \ref{fig:resultsBactClass} respectively for the digit and microbiological image experiments. In Figures \ref{fig:databaseDigitClass} and \ref{fig:databaseBactClass}, a data set image from each class is shown. Some typical representative patterns computed with JPATS, PATS  and the reference methods are shown in Figures \ref{fig:classDigitsJPATS}-\ref{fig:classDigitsSAS} and \ref{fig:classBactJPATS}-\ref{fig:classBactSAS}. Figures  \ref{fig:errorDigitClass} and \ref{fig:errorBactClass} show the misclassification rates of test images (in percentage) vs. the number of atoms per class. Both plots are obtained by averaging the results of 5 repetitions of the experiment with different training and test sets. The results show that JPATS  yields the best classification performance in general. \revis{Figures \ref{fig:resultsDigitClass} and \ref{fig:resultsBactClass} suggest that  JPATS has better classification performance although PATS produces visually more pleasant patterns. This can be explained as follows. PATS is designed to minimize the approximation error; and the assessment of the visual quality of the computed patterns is rather dependent on their approximation capabilities. The local features that are common to different classes appear in the representative patterns of all these classes built with PATS, which produces an output that matches visual perception. However, if a local feature is common to several classes, its inclusion in the representative patterns does not contribute much to the discrimination among classes; therefore, these non-distinctive features are not emphasized in the output of JPATS. On the other hand, the local features that are rather special to one class are more pronounced in JPATS compared to PATS. In fact, due to the classification error term in JPATS, the algorithm tends to select atoms that ``push'' a manifold away from the samples of other classes.}\footnote{\revis{For instance, in Figure \ref{fig:classDigitsJPATS}, the top and bottom ``arcs'' of the ``8'' digit are not as apparent, since the other digits also have similar features (all other digits have the top part; and ``2'', ``3'' and ``5'' have a bottom part). However, the crossover of  ``8'' is specific to this class; therefore, it is prominent in the output. Similarly, the straight edge of ``9'' is also characteristic of this class and emphasized in the learned pattern.}}

\begin{figure}
 \centering
 \includegraphics[width=8cm]{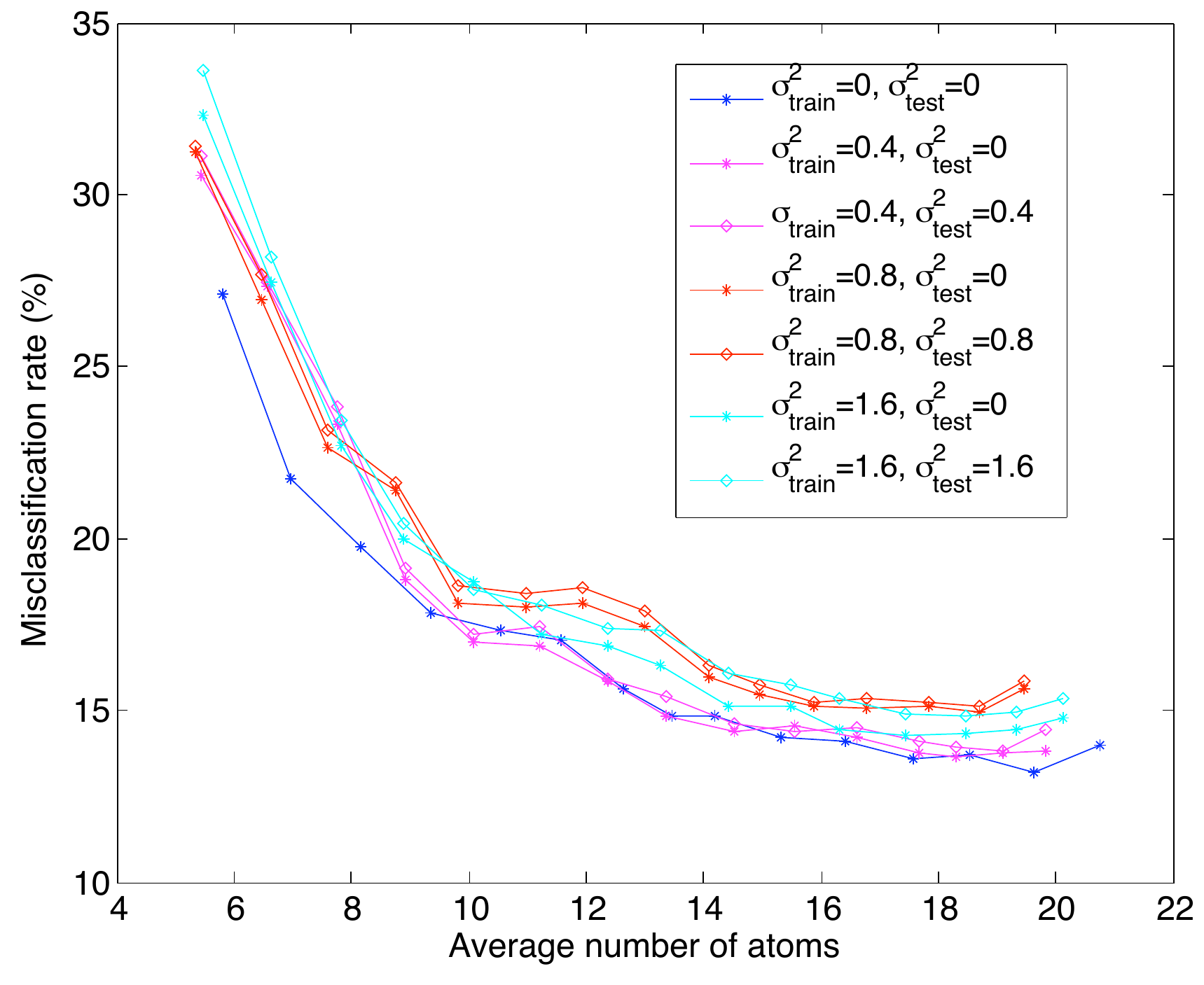} 
  \caption{Performance of JPATS on noisy data. The noise variance $\sigma^2=1.6$ corresponds to an SNR of 6.35 dB.}
    \label{fig:resultsDigitClassNoisy}
\end{figure}

\revis{Next, we examine the effect of data noise on the performance of JPATS. We create several data sets by corrupting the digits data set used in the previous experiment with additive Gaussian noise of different variances. For each noise level, we look into two cases, where only training images are corrupted in the first one, and both training and test images are corrupted in the second one. The misclassification rate of test images are plotted in Figure \ref{fig:resultsDigitClassNoisy}, where $\sigma_{train}^2$ and $\sigma_{test}^2$ denote the noise variances of training and test images. The data noise has a small influence on the performance of the algorithm. The final increase in the misclassification rate is bounded by $2.7\%$ even when the noise energy reaches $23\%$ of the signal energy. The robustness to noise is achieved due to the fact that the algorithm is designed to generate a smooth pattern that fits all images simultaneously, which enables it to smooth data noise. The other PTM-based methods are also expected to exhibit similar noise behaviors.} 

\begin{figure}
 \centering
 \includegraphics[width=8cm]{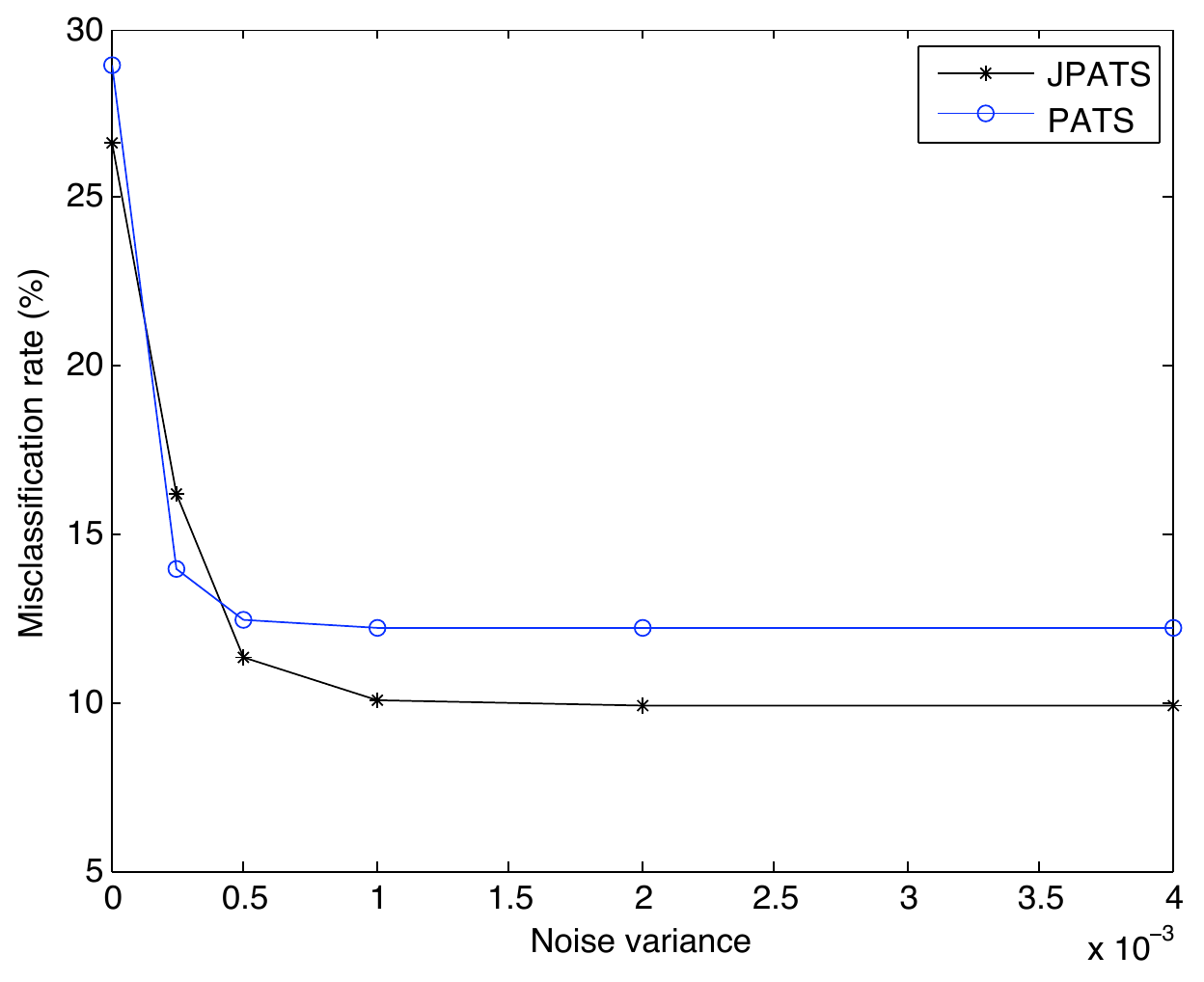} 
  \caption{Performance of PATS and JPATS in a classification setting with outlier test images that do not belong to any class. For the noise variance $\sigma^2=0.5 \times 10^{-3}$, the ratio between the norms of the noise component and the average component of an outlier image is $0.65$.}
    \label{fig:resultsDigitClassOutliers}
\end{figure}

\revistwo{Finally, we evaluate the performances of  PATS and JPATS in a setting where the test images contain some outliers that do not belong to any of the classes. We run the experiment on the same digit data set that has been used in the previous experiment of Figure \ref{fig:resultsDigitClass}. The training phase of the algorithms is as before: In both methods, the manifolds are learned using only training images of known classes. However, test images are contaminated with 200 outlier images that do not belong to any of the target classes, where the number of test images in each class is also 200. Each outlier image is generated by randomly selecting one test image from each class, taking the average of these images, corrupting the average image with additive Gaussian noise, and finally normalizing it. Thus, all outlier images have unit norm, while a typical class-sample test image with unit scale ($s_x=1, s_y=1$) also has unit norm. Then, test images are classified using the manifolds learned with PATS and JPATS as follows. If the distance between a test image and the closest manifold is larger than a threshold, the image is labeled as an ``outlier''; and if this distance is smaller than the threshold, it is assigned the class label of the closest manifold as before. The experiment is repeated for different values of the noise variance for the Gaussian noise component of the outlier images. The threshold used for each noise level is numerically selected in a sample run of the experiment such that it gives the best classification rate and fixed for the other runs, separately for PATS and JPATS. The results are presented in Figure \ref{fig:resultsDigitClassOutliers}, which are the average of 5 runs. The misclassification rate is the percentage of test images that have not been assigned the correct class label or the correct ``outlier'' label. In the plots shown in Figure \ref{fig:resultsDigitClassOutliers}, the noise variance $0$ corresponds to the case that outlier images are the averages of some test images coming from different classes. This is the most challenging instance of the experiment, as outliers come from a region close to class samples and it is thus difficult to distinguish them from class samples. Consequently, the optimal threshold that gives the smallest misclassification rate is high for this instance, resulting in labeling all outliers as class samples. The performance of JPATS is better than PATS in this case, as the overall classification rate is determined by the classification rate of class samples. Then, as the noise variance is increased, the components of the outlier images in random directions in the image space get more dominant, making it thus easier to distinguish them from class samples. It is seen that JPATS performs better than PATS in most cases. However, for the smallest nonzero noise variance value $\sigma^2=0.25\times 10^{-3}$, PATS is observed to give a better classification rate than JPATS. This can be explained as follows. In this experiment, JPATS is trained according to the hypothesis that all test images belong to a valid class. For this reason, in order to increase the distance between a manifold and the samples of other classes, JPATS may occasionally pick some atoms that push a manifold away from the samples of its own class as well. This slight increase in the distance between the manifold and the samples of its own class renders it difficult for JPATS to distinguish between real class samples and challenging outliers that are very close to class samples. In order to get the best performance from JPATS in such a setting with outliers, one can tune the $\alpha$ parameter to a suitable value depending on outlier characteristics such that a sufficiently strict control is imposed on the distance between the learned manifolds and the samples of their own classes through the approximation term $E_a$.}

\section{Complexity Analysis}
\label{sec:complexity}

Let us now examine the complexities of the proposed algorithms. We begin with the PATS method summarized in Algorithm \ref{alg:pats}. There are three blocks in the main loop of the algorithm. The first one minimizes $\tilde{E}$ with DC programming, the second one minimizes $\hat E$ via gradient descent, and the third one computes the projections of $\{ u_i\}$ on the manifold. In the analysis of the first block, it is important to distinguish between the complexities of the DC solver and the computation of the DC decomposition of $\tilde{E}$. The former depends on the selected solver. The cutting plane method involves the construction of polytopes in the search space; therefore, it has an exponential complexity in the number of atom parameters $s$ (dimension of $\mathcal{D}$). \revistwo{One iteration of the cutting plane algorithm typically takes a few seconds in a MATLAB implementation, and around 40-60 iterations are done for atom selection.} However, one can also use a technique such as DCA \cite{Tao97}. In this case, the solution of the dual convex programs involves the evaluation of the subdifferentials of the functions constituting the DC decomposition. In our problem, this corresponds to the evaluation of gradients since the decomposing functions are differentiable. The gradient can be numerically evaluated using finite differences; therefore, the complexity of such a solver is at most linear in $s$. Next, it can be seen from  (\ref{eq:DCnorm2hate}) and (\ref{eq:appTildeEfin}) that the cost of computing the DC decomposition of $\tilde{E}$ is linear in the image resolution $n$ and the number of samples $N$. Hence, the complexity of the first block becomes $O(2^s \, n \, N)$ for cutting plane, and $O(s \, n \, N)$ for DCA. In the analysis of the second block, \revistwo{it can be easily shown that the complexity of calculating the vector $w_i  - T_{i}  (T_{i}^{\mbox{\scriptsize{T}}} \, T_{i})^{-1} \, T_{i}^{\mbox{\scriptsize{T}}} w_i $ in (\ref{eq:hatEfinal}) is $O( d\, n^2)$, where $d$ is the dimension of $\mathcal{M}(p)$. Therefore, the cost of computing the total squared tangent distance $\hat E$ in (\ref{eq:hatEfinal}) is obtained as $O(d \, n^2 \, N)$.} As we minimize $\hat E$ with gradient descent using finite differences, the complexity of the second block is $O(s \, d \, n^2 \, N )$. Finally, the cost of updating the projections of $\{ u_i\}$ on $\mathcal{M}(p)$ is  $O(d \, n \, N)$ in our actual implementation, because we minimize the distance between each $\{ u_i\}$ and $\mathcal{M}(p)$ by performing a line search along each dimension of $\mathcal{M}(p)$. Thus, taking DCA as a reference for the first block, we can summarize the overall complexity of PATS as $O(s \, d \, n^2 \, N )$.

We examine the complexity of JPATS given in Algorithm \ref{alg:jpats} similarly. From (\ref{eq:appTildeE}) it is seen that the cost of computing the DC decomposition of $\tilde E$ is linear in $N_J$ and $n$, where $N_J = \sum_{m=1}^M N_m =\sum_{m=1}^M | R^m |$. The complexity of the first block with respect to DCA is therefore $O(s \, n \, N_J)$. Then, (\ref{eq:hatEa}) and (\ref{eq:hatEpsilon2}) show that the cost of computing $\hat E$ is $O(d \, n^2 \, N_J)$. The complexity of the second block is thus $O(s \, d \, n^2 \, N_J)$. Finally, the third block has complexity $O(d \, n \, N_J \, M)$, since each image is reprojected on each manifold. Therefore, the overall complexity of JPATS is $O\big( N_J (s\, d \, n^2 + d \, n \, M) \big) $. \revis{The complexity of selecting an atom with ``SMP on aligned patterns'', which has the closest performance to the proposed methods, can be similarly obtained as $O(N_J \, n \, D)$, where $D$ denotes the cardinality of the discrete dictionary used.} We remark that the proposed method is more suitable for applications where the manifolds are learned ``offline'' and then used for the classification of test data. Moreover, there might be ways to improve the complexity-accuracy tradeoff depending on the application. For instance, one might prefer to sacrifice on accuracy for a less complex solution by omitting step \ref{state:atomSelectJPATS} or \ref{state:gradientDesJPATS} of Algorithm \ref{alg:jpats}. Also, if the class-representative manifolds are well-separated, it may be sufficient to use the PATS algorithm instead of JPATS. \revis{An option for achieving a high-speed PTM learning is to build a tentative representative pattern, for instance with ``SMP on aligned patterns'', in a preliminary analysis step and register the input images with respect to this pattern. Then, one may speed up the learning significantly by discarding the projection update steps and optimizing the atoms of the representative pattern by minimizing only the error in (\ref{eq:tildeE}) with a fast minimizer such as the gradient descent algorithm.}

\section{Conclusion}
\label{sec:Conclusion}
We have studied the problem of building smooth pattern transformation manifolds for the transformation-invariant representation of sets of visual signals. The manifold learning problem is cast as the construction of a representative pattern as a linear combination of smooth parametric atoms. The manifold is then created by geometric transformations of this pattern. The smoothness of the computed manifolds is ensured by the smoothness of the constituting parametric atoms. We have described a single manifold learning algorithm for approximation and a multiple manifold learning algorithm for classification. Experimental results show that the proposed methods provide a good approximation and classification accuracy compared to reference methods. \revis{A future direction to explore is the amelioration of the sensitivity of the methods to the initialization of projection parameters.} The presented methods are applicable to unregistered data that can be approximated by 2-D pattern transformations with a known transformation model. Our study can find several applications in the transformation-invariant representation, registration, coding and classification of images.

\section{Acknowledgment}
\label{sec:acknowledge}
The authors would like to thank Dr. E. Kokiopoulou for providing an implementation of the cutting plane algorithm and the helpful discussions on DC optimization.

\ifCLASSOPTIONcaptionsoff
  \newpage
\fi


\bibliographystyle{IEEEtran}
\bibliography{refs}

\begin{thebibliography}{10}
\providecommand{\url}[1]{#1}
\csname url@samestyle\endcsname
\providecommand{\newblock}{\relax}
\providecommand{\bibinfo}[2]{#2}
\providecommand{\BIBentrySTDinterwordspacing}{\spaceskip=0pt\relax}
\providecommand{\BIBentryALTinterwordstretchfactor}{4}
\providecommand{\BIBentryALTinterwordspacing}{\spaceskip=\fontdimen2\font plus
\BIBentryALTinterwordstretchfactor\fontdimen3\font minus
  \fontdimen4\font\relax}
\providecommand{\BIBforeignlanguage}[2]{{%
\expandafter\ifx\csname l@#1\endcsname\relax
\typeout{** WARNING: IEEEtran.bst: No hyphenation pattern has been}%
\typeout{** loaded for the language `#1'. Using the pattern for}%
\typeout{** the default language instead.}%
\else
\language=\csname l@#1\endcsname
\fi
#2}}
\providecommand{\BIBdecl}{\relax}
\BIBdecl

\bibitem{1028585}
S.~G. Mallat and Z.~Zhang, ``Matching pursuits with time-frequency
  dictionaries,'' \emph{IEEE Trans. Signal Process.}, vol.~41, no.~12, pp.
  3397--3415, Dec 1993.

\bibitem{Vural2011learn}
E.~{V}ural and P.~Frossard, ``Learning pattern transformation manifolds with
  parametric atom selection,'' in \emph{Proc. Sampling Theory and
  Applications}, May 2011.

\bibitem{efiDC}
E.~Kokiopoulou and P.~Frossard, ``Minimum distance between pattern
  transformation manifolds: Algorithm and applications,'' \emph{IEEE Trans.
  Pattern Anal. Mach. Intell.}, vol.~31, no.~7, pp. 1225--1238, Jul. 2009.

\bibitem{266187}
J.~B. Tenenbaum, V.~de~Silva, and J.~C. Langford, ``A global geometric
  framework for nonlinear dimensionality reduction.'' \emph{Science}, vol. 290,
  no. 5500, pp. 2319--2323, Dec. 2000.

\bibitem{Roweis00nonlineardimensionality}
S.~T. Roweis and L.~K. Saul, ``Nonlinear dimensionality reduction by locally
  linear embedding,'' \emph{Science}, vol. 290, pp. 2323--2326, Dec. 2000.

\bibitem{donoho03hessian}
D.~L. Donoho and C.~Grimes, ``{Hessian eigenmaps: Locally linear embedding
  techniques for high-dimensional data},'' in \emph{Proc. Natl. Acad. Sci.
  USA}, vol. 100, no.~10, May 2003, pp. 5591--5596.

\bibitem{Bengio04out}
Y.~Bengio, J.~F. Paiement, P.~Vincent, O.~Delalleau, N.~L. Roux, and M.~Ouimet,
  ``Out-of-sample extensions for {LLE}, {ISOMAP}, {MDS}, {E}igenmaps, and
  {S}pectral {C}lustering,'' in \emph{Adv. Neural Inf. Process. Syst.}\hskip
  1em plus 0.5em minus 0.4em\relax MIT Press, 2004, pp. 177--184.

\bibitem{DollarRabaudBelongieICML07manifold}
P.~Doll\'ar, V.~Rabaud, and S.~Belongie, ``Non-isometric manifold learning:
  Analysis and an algorithm,'' in \emph{Int. Conf. Mach. Learn.}, June 2007.

\bibitem{AlvarezMezaVDAC11}
A.~M. {\'A}lvarez-Meza, J.~Valencia-Aguirre, G.~Daza-Santacoloma, C.~D.
  Acosta-Medina, and G.~Castellanos-Dom\'{\i}nguez, ``Image synthesis based on
  manifold learning,'' in \emph{CAIP (2)}, 2011, pp. 405--412.

\bibitem{deridder:supervised}
D.~de~Ridder, O.~Kouropteva, O.~Okun, M.~Pietikainen, and R.~P.~W. Duin,
  ``Supervised locally linear embedding.'' in \emph{Proc. Int. Conf. Art. Neur.
  Networks}, 2003, pp. 333--341.

\bibitem{Miller2006}
E.~Learned-Miller, ``Data driven image models through continuous joint
  alignment,'' \emph{{IEEE} Trans. on Pattern Anal. and Machine Intel.},
  vol.~28, pp. 236--250, 2006.

\bibitem{atkeson_locally1}
C.~Atkeson, A.~W. Moore, and S.~Schaal, ``Locally weighted learning,'' \emph{AI
  Review}, vol.~11, pp. 11--73, April 1997.

\bibitem{MooreSD97}
A.~W. Moore, J.~Schneider, and K.~Deng, ``Efficient locally weighted polynomial
  regression predictions,'' in \emph{Int. Conf. on Machine Learning}, 1997, pp.
  236--244.

\bibitem{Jonsson2007}
E.~Jonsson and M.~Felsberg, ``Accurate interpolation in appearance-based pose
  estimation,'' in \emph{Proc. 15th Scandinavian Conf. on Image Anal.}, ser.
  SCIA'07.\hskip 1em plus 0.5em minus 0.4em\relax Berlin, Heidelberg:
  Springer-Verlag, 2007, pp. 1--10.

\bibitem{1140735}
J.~A. Tropp, A.~C. Gilbert, and M.~J. Strauss, ``Algorithms for simultaneous
  sparse approximation part {I}: Greedy pursuit,'' \emph{Signal Processing},
  vol.~86, no.~3, pp. 572--588, 2006.

\bibitem{EPFLARTICLE91054}
E.~Kokiopoulou and P.~Frossard, ``Semantic coding by supervised dimensionality
  reduction,'' \emph{{IEEE} {T}rans. {M}ultimedia}, vol.~10, no.~5, pp.
  806--818, 2008.

\bibitem{Mailhe08shift}
B.~Mailh\'{e}, S.~Lesage, R.~Gribonval, F.~Bimbot, and P.~Vandergheynst,
  ``Shift-invariant dictionary learning for sparse representations: {E}xtending
  {K-SVD},'' in \emph{Proc. Eur. Sig. Proc. Conf.}, 2008.

\bibitem{Jost2006_1450}
P.~Jost, S.~Lesage, P.~Vandergheynst, and R.~Gribonval, ``{MoTIF}: An efficient
  algorithm for learning translation-invariant dictionaries,'' in \emph{Proc.
  IEEE Int. Conf. Acoust., Speech, Signal Proc.}, vol.~5, 2006, pp. 857--860.

\bibitem{Mairal2007}
J.~Mairal, G.~Sapiro, and M.~Elad, ``Multiscale sparse image representation
  with learned dictionaries,'' in \emph{Proc. IEEE Int. Conf. Image Proc.},
  vol.~3, Sep 2007, pp. 105--108.

\bibitem{SallOls2002}
P.~Sallee and B.~Olshausen, ``Learning sparse multiscale image
  representations,'' in \emph{Adv. in Neur. Inf. Proc. Sys.}\hskip 1em plus
  0.5em minus 0.4em\relax MIT Press, 2002.

\bibitem{Ekanad2011}
C.~Ekanadham, D.~Tranchina, and E.~P. Simoncelli, ``Sparse decomposition of
  transformation-invariant signals with continuous basis pursuit,'' in
  \emph{Proc. IEEE Int. Conf. Acoust., Speech, Signal Proc.}, 2011, pp.
  4060--4063.

\bibitem{Antoine2004}
J.~Antoine, R.~Murenzi, P.~Vandergheynst, and S.~Ali, \emph{Two-{D}imensional
  {W}avelets and their {R}elatives}, ser. Signal Processing.\hskip 1em plus
  0.5em minus 0.4em\relax Cambridge University Press, 2004.

\bibitem{introGlobOpt}
R.~Horst, P.~M. Pardalos, and N.~V. Thoai, \emph{Introduction to Global
  Optimization}.\hskip 1em plus 0.5em minus 0.4em\relax Dordrecht, The
  Netherlands: Kluwer Academic Publishers, 2000.

\bibitem{Hartman1959}
P.~Hartman, ``On functions representable as a difference of convex functions,''
  \emph{Pacific Journal of Math.}, no.~9, pp. 707--713, 1959.

\bibitem{Horst1999}
R.~Horst and N.~V. Thoai, ``{DC} programming: overview,'' \emph{J. Optim.
  Theory Appl.}, vol. 103, pp. 1--43, October 1999.

\bibitem{Tao97}
P.~Tao and L.~An, ``Convex analysis approach to {DC} programming: Theory,
  algorithms and applications,'' \emph{Acta Mathematica Vietnamica}, vol.~22,
  no.~1, pp. 289--355, 1997.

\bibitem{Yuille02theconcave}
A.~Yuille, A.~Rangarajan, and A.~L. Yuille, ``The concave-convex procedure
  {(CCCP)},'' in \emph{Adv. in Neur. Inf. Proc. Sys.}\hskip 1em plus 0.5em
  minus 0.4em\relax MIT Press, 2002.

\bibitem{668381}
P.~Simard, Y.~LeCun, J.~S. Denker, and B.~Victorri, ``Transformation invariance
  in pattern recognition-tangent distance and tangent propagation,'' in
  \emph{Neural Networks: Tricks of the Trade}.\hskip 1em plus 0.5em minus
  0.4em\relax New York: Springer-Verlag, 1998.

\bibitem{Vural2011appx}
E.~Vural and P.~Frossard, ``Approximation of pattern transformation manifolds
  with parametric dictionaries,'' in \emph{Proc. IEEE Int. Conf. Acoust.,
  Speech, Signal Process}, May 2011, pp. 977--980.

\bibitem{lecun98}
Y.~LeCun, L.~Bottou, Y.~Bengio, and P.~Haffner, ``Gradient-based learning
  applied to document recognition,'' \emph{Proceedings of the IEEE}, vol.~86,
  no.~11, pp. 2278--2324, Nov. 1998.

\bibitem{CKFaceDatabase}
T.~Kanade, J.~F. Cohn, and Y.~Tian, ``Comprehensive database for facial
  expression analysis,'' \emph{Fourth IEEE International Conference on
  Automatic Face and Gesture Recognition}, pp. 46--53, 2000.

\bibitem{SalaLlonch2010}
R.~Sala~Llonch, E.~Kokiopoulou, I.~To\v{s}i\'{c}, and P.~Frossard, ``3d face
  recognition with sparse spherical representations,'' \emph{Pattern Recogn.},
  vol.~43, no.~3, pp. 824--834, Mar. 2010.

\bibitem{KCLee05}
K.~Lee, J.~Ho, and D.~Kriegman, ``Acquiring linear subspaces for face
  recognition under variable lighting,'' \emph{IEEE Trans. Pattern Anal. Mach.
  Intelligence}, vol.~27, no.~5, pp. 684--698, 2005.

\bibitem{vuralManifDisc}
E.~Vural and P.~Frossard, ``Discretization of {P}arametrizable {S}ignal
  {M}anifolds,'' \emph{{IEEE} {T}ransactions on {I}mage {P}rocessing}, vol.~20,
  no.~12, pp. 3621--3633, 2011.

\bibitem{NHMpage}
\BIBentryALTinterwordspacing
{Natural History Museum}. (2011) Microbiology video collection. [Online].
  Available:
  \url{http://www.nhm.ac.uk/research-curation/research/projects/protistvideo/a%
bout.dsml}
\BIBentrySTDinterwordspacing

\bibitem{munkres1991analysis}
J.~Munkres, \emph{Analysis on manifolds}, ser. Advanced Book Classics.\hskip
  1em plus 0.5em minus 0.4em\relax Addison-Wesley Pub. Co., Advanced Book
  Program, 1991.

\end{thebibliography}

%
%
%

%
%
%

\appendix
\maketitle{\textit{A. Proof of Proposition \ref{prop:DCformTildeE} }}

Before showing the DC property of the objective function, we list below some useful properties of DC functions from \cite{introGlobOpt} and \cite{efiDC}.\\

\begin{proposition}
\label{prop:DC}
\begin{enumerate}[(a)]
\item Let $\{f_i\}_{i=1}^m$ be a set of DC functions with decompositions $f_i=g_i-h_i$ and let $\{ \lambda_i \}_{i=1}^m \subset \mathbb{R}$. Then $\sum_{i=1}^m \lambda_i f_i$ has the following DC decomposition \cite{introGlobOpt}, \cite{efiDC}.  \label{prop:DCa}

\begin{eqnarray*}
\begin{split}
\sum_{i=1}^m \lambda_i f_i &= \left(  \sum_{i: \lambda_i >0} \lambda_i g_i  - \sum_{i: \lambda_i <0} \lambda_i h_i    \right) \\
&-   \left(  \sum_{i: \lambda_i >0} \lambda_i h_i  - \sum_{i: \lambda_i <0} \lambda_i g_i    \right)
\end{split}
\end{eqnarray*}\\

\item Let $f_1=g_1 - h_1$ and $f_2=g_2 - h_2$ be DC functions with nonnegative convex parts $g_1, h_1, g_2, h_2$. Then the product $f_1 \, f_2$ has the DC decomposition
\begin{eqnarray*}
\begin{split}
f_1 f_2 &= \frac{1}{2} \left(  (g_1+g_2)^2 + (h_1 + h_2)^2  \right) \\
		&-\frac{1}{2} \left(  (g_1+h_2)^2 + (g_2 + h_1)^2  \right), 
\end{split}
\end{eqnarray*}
which has nonnegative convex parts  \cite{introGlobOpt}, \cite{efiDC}.\\
\label{prop:DCc}
\end{enumerate}
\end{proposition}

Now we can give a proof of Proposition \ref{prop:DCformTildeE}.\\

\begin{proof}
Let $  \tilde e =  v -  c \, U_{\lambda} (\phi_{\gamma} )$ denote the difference vector between an image $v$ and an atom  $\phi_{\gamma}$ transformed by $\lambda$ with a coefficient $c$. We first show that the components (pixels) of $U_{\lambda} (\phi_{\gamma} )$ are DC functions of $\gamma$ and $c$. Remember that $U_{\lambda} (\phi_{\gamma} )$ has been defined as a discretization of $A_{\lambda}(\phi_{\gamma})= A_{\lambda}(B_{\gamma} (\phi))$. We can write 
\[ A_{\lambda}(B_{\gamma} (\phi))(x,y) = B_{\gamma} (\phi)(x',y')  = \phi(\tilde x, \tilde y) \]
where all coordinate variables are related by
\[  (\tilde x, \tilde y) = b_{\gamma}(x',y') =  (b_{\gamma} \circ a_{\lambda}) ( x, y) .\]
Then, we get
\[ A_{\lambda}(B_{\gamma} (\phi))(x,y) = \phi( b_{\gamma}( a_{\lambda}( x, y) ) )  .\]
The $l$-th component $U_{\lambda} (\phi_{\gamma} )(l)$ of $U_{\lambda} (\phi_{\gamma} )$ corresponds to a certain point with coordinates $(x,y)$ such that 
\[ U_{\lambda} (\phi_{\gamma} )(l)=A_{\lambda}(B_{\gamma} (\phi))(x,y) =  \phi( b_{\gamma}( a_{\lambda}( x, y) ) ).\]
Here, $b$ is a smooth function of $\gamma$, and $\phi$ is also a smooth function. Therefore, being a composition of two smooth functions, $U_{\lambda} (\phi_{\gamma} )(l)$ is smooth as well (\cite{munkres1991analysis}, Corollary 7.2), and thus DC by Proposition \ref{prop:smoothIsDC}.

In the following, we show the DC property of $\tilde E$. We describe at the same time a procedure to compute the DC decomposition of $\tilde E$ if a DC decomposition of $U_{\lambda} (\phi_{\gamma} )(l)$ is available. We expand $\| \tilde e \|^2$ in terms of the errors at individual pixels as
\begin{equation}
\begin{split}
\|  \tilde e \|^2 &= \| v -  c \,  U_{\lambda} (\phi_{\gamma} ) \|^2 \\
                         &= \sum_{l=1}^n \big( v^2(l) -2 \, v(l) \, c  \, U_{\lambda} (\phi_{\gamma} ) (l) + c^2 \, U_{\lambda}^2 (\phi_{\gamma} )  (l) \big)
\end{split}
\label{eq:DCnorm2hate}
\end{equation}
where $v(l)$ is the $l$-th component of $v$. The term $v(l)$ is constant with respect to $\gamma$ and $c$. Using the DC decomposition of $ U_{\lambda} (\phi_{\gamma} ) (l) $ and decomposing $c$ as $c=0.5 \, (c+1)^2-0.5\, (c^2+1)$, we can compute the DC decomposition of the second term $-2 \, v(l) \, c  \, U_{\lambda} (\phi_{\gamma} ) (l) $ from Propositions \ref{prop:DC}.\ref{prop:DCc} and \ref{prop:DC}.\ref{prop:DCa}. One can also obtain the decomposition of the last term $c^2 \, U_{\lambda}^2 (\phi_{\gamma} )  (l) $ by applying Proposition \ref{prop:DC}.\ref{prop:DCc}. Finally, the DC decompositions of $\| \tilde e \|^2$ and 
\begin{equation}
\label{eq:appTildeEfin}
\tilde E= \sum_{i=1}^N  \| \tilde e_i \|^2
\end{equation}
simply follow from Proposition \ref{prop:DC}.\ref{prop:DCa}.\\
 \end{proof}

%
%
%
%
%
\maketitle{\textit{B. Derivation of total squared tangent distance $ \hat{E}$ }}\\

The first order approximation of the manifold $\mathcal{M}( p_j)  $ around the point $U_{\lambda_i}(p_j)$ is given by \cite{Vural2011appx}
\begin{equation*}
\mathcal{M}(p_j)  \approx  \mathcal{S}_i (p_j) = \{ U_{\lambda_i}( p_j) + T_{i} \, z: z \in \mathbb{R}^{d\times 1}  \}
\end{equation*}
where $T_{i}$ is an $n \times d$ matrix consisting of tangent vectors. The $k$-th column of $T_{i}$ is the tangent vector $ \partial / \partial_k \, U_{\lambda_i}( p_j )$, which is the derivative of the manifold point $U_{\lambda_i}(p_j)$ with respect to the $k$-th transformation parameter $\lambda_i(k)$. 
The orthogonal projection of $u_i$ on $\mathcal{S}_i( p_j )$ is given by $\hat{u}_i = U_{\lambda_i}(p_j)  + T_{i} z^{*}$, where the coefficient vector $z^{*}$ of the projection is $z^{*} = (T_{i}^{\mbox{\scriptsize{T}}} \, T_{i})^{-1} \, T_{i}^{\mbox{\scriptsize{T}}} (u_i - U_{\lambda_i}(p_j ) )$. Hence, the difference vector $\hat{e}_i$ between $u_i$ and $\hat{u}_i$ is
\begin{eqnarray*}
\begin{split}
\hat{e}_i &= u_i - \hat{u}_i \\
       & = u_i - U_{\lambda_i}(p_j )  - T_{i}  (T_{i}^{\mbox{\scriptsize{T}}} \, T_{i})^{-1} \, T_{i}^{\mbox{\scriptsize{T}}} (u_i - U_{\lambda_i}(p_j) ). 
\end{split}
\end{eqnarray*}
Letting $ w_i = u_i - U_{\lambda_i}(p_{j}) $, we get $\hat{E}$ as
\begin{equation}
\label{eq:hatEfinal}
\hat{E} = \sum_{i=1}^N  \| w_i  - T_{i}  (T_{i}^{\mbox{\scriptsize{T}}} \, T_{i})^{-1} \, T_{i}^{\mbox{\scriptsize{T}}} w_i   \|^2.
\end{equation}
\\

%
%
%
%
%
%
\maketitle{\textit{C. Computation of the DC Decompositions in Section \ref{ssec:appxExperiments}}}\\

In the derivation of the DC decompositions of $U_{\lambda} (\phi_{\gamma} ) (l)$ and $\tilde E$, we build on the results from \cite{efiDC}, where a DC decomposition of the distance between a query pattern and the 4-dimensional transformation manifold of a reference pattern is derived. We first give the following results from \cite{efiDC}.\\

\begin{proposition}
\label{prop:DCfromEfi}

\begin{enumerate}[(a)]

\item Let $f: \mathbb{R}^s \rightarrow \mathbb{R}$ be a DC function with decomposition $f(x) = g(x) - h(x)$ and $q: \mathbb{R} \rightarrow \mathbb{R}$ be a convex function. Then $q(f(x))$ is DC and has the decomposition \cite{efiDC}  \label{prop:DCb}

\begin{equation*}
q(f(x))= p(x) - K [g(x) + h(x)],
\end{equation*}

where $p(x) = q(f(x)) + K [g(x)+h(x)]$ is a convex function and $K$ is a constant satisfying $K>|q'(f(x))|$.\\

\item Let $\psi \in [0, 2\pi)$ and $\sigma \in \mathbb{R}^{+}$. Then the following functions have DC decompositions with nonnegative convex parts: $\cos(\psi), \, \sin(\psi), \frac{\cos(\psi)}{\sigma},\frac{ \sin(\psi)}{\sigma}$. See \cite{efiDC} for computation details.  \label{prop:DCd}\\
\end{enumerate}
\end{proposition}

The relation between the transformed atom and the mother Gaussian function is given by the change of variables 
\[ A_{\lambda} (\phi_{\gamma} ) (x,y) = \phi(\tilde x, \tilde y) = \sqrt \frac{2}{\pi} e^{-(\tilde{x}^2+\tilde{y}^2)} .\]
Let the $l$-th pixel of $U_{\lambda}(\phi_{\gamma})$ correspond to the coordinates $(x,y)$ in $A_{\lambda} (\phi_{\gamma} ) $ and the coordinates $(\tilde x, \tilde y)$ in $\phi$. From (\ref{eq:TransModel}) and (\ref{eq:DictModel}), $(\tilde x, \tilde y)$ can be derived in the form
\begin{eqnarray*}
\begin{split}
\tilde x &= \, \nu \frac{\cos(\psi)}{\sigma_x} + \xi \frac{\sin(\psi)}{\sigma_x}
              - \frac{ \cos(\psi) \tau_x }{\sigma_x} - \frac{\sin(\psi) \tau_y}{\sigma_x}\\
\tilde y &= \, \xi \frac{\cos(\psi)}{\sigma_y} - \nu \frac{\sin(\psi)}{\sigma_y} 
             - \frac{\cos(\psi) \tau_y }{\sigma_y} + \frac{\sin(\psi) \tau_x}{\sigma_y}
\end{split}
\end{eqnarray*}
where 
\begin{eqnarray*}
\begin{split}
\nu &= \frac{\cos(\theta) x  +   \sin(\theta) y  -  \cos(\theta) t_x - \sin(\theta) t_y }{s_x}\\
\xi &= \frac{-\sin(\theta) x  +   \cos(\theta) y  +  \sin(\theta) t_x - \cos(\theta) t_y }{s_y}.
\end{split}
\end{eqnarray*}
Here $\nu$ and $\xi$ are functions of the transformation parameters $\lambda$ and the coordinates $(x,y)$ but they stay constant with respect to the atom parameters $\gamma$. Now we explain how the coordinate variables $\tilde x$ and $\tilde y$ can be expanded in the DC form in terms of the atom parameter variables. 

Firstly, from Proposition \ref{prop:DCfromEfi}.\ref{prop:DCd}, notice that the decompositions of the functions $\{ \frac{\cos(\psi)}{\sigma_x}, \frac{\sin(\psi)}{\sigma_x}, \frac{\cos(\psi)}{\sigma_y},  \frac{\sin(\psi)}{\sigma_y}  \}$ can be computed as explained in \cite{efiDC}. The first two terms in $\tilde x$ and $\tilde y$ are given by the product of one of these functions with a constant term ($\nu$ or $\xi$). Therefore, we can get the decompositions of these terms using Proposition \ref{prop:DC}.\ref{prop:DCa}.

Then, observe that $\tau_x$ has a decomposition as $\tau_x=0.5\, (\tau_x+1)^2 - 0.5 \, (\tau_x^2 +1)$ \cite{efiDC} and the decomposition of $\tau_y$ is obtained in the same manner. Thus, one can obtain the DC decompositions of the last two terms $\{ \frac{ \cos(\psi) \tau_x }{\sigma_x}, \frac{\sin(\psi) \tau_y}{\sigma_x},   \frac{\cos(\psi) \tau_y }{\sigma_y},  \frac{\sin(\psi) \tau_x}{\sigma_y} \}$ in $\tilde x$ and $\tilde y$ by applying the product property in Proposition \ref{prop:DC}.\ref{prop:DCc} on the decompositions of the terms in $\{ \frac{\cos(\psi)}{\sigma_x}, \frac{\sin(\psi)}{\sigma_x}, \frac{\cos(\psi)}{\sigma_y},  \frac{\sin(\psi)}{\sigma_y}  \}$ and $ \{ \tau_x, \tau_y \}$. Hence, having computed the decompositions of all additive terms in $\tilde x$ and $\tilde y$, one can obtain the decompositions of $\tilde x$ and $\tilde y$ using Proposition \ref{prop:DC}.\ref{prop:DCa}. 

Let $\tilde z= \tilde{x}^2 + \tilde{y}^2$. The decompositions of $ \tilde{x}^2$ and $ \tilde{y}^2 $ can be obtained by applying the product property in Proposition \ref{prop:DC}.\ref{prop:DCc} on the decompositions of $\tilde x$ and $\tilde y$. Then, the decomposition of $\tilde z$ follows from Proposition \ref{prop:DC}.\ref{prop:DCa}. Expressing the mother function 
\[
\phi(\tilde x, \tilde y)=  \sqrt \frac{2}{ \pi} e^{- \tilde z}
\]
 as a convex function of $\tilde z$, Proposition \ref{prop:DCfromEfi}.\ref{prop:DCb} provides the decomposition of $\phi(\tilde x, \tilde y)$. Thus, we obtain the decomposition of $U_{\lambda} (\phi_{\gamma} ) (l)$.

After this point, the decomposition of $\tilde E$ can be computed based on the description in Appendix B. However, notice that for this special case of Gaussian mother function, the decomposition of the term $U^2_{\lambda}(\phi_{\gamma})(l)$ in (\ref{eq:DCnorm2hate}) can also be obtained by writing $A_{\lambda}^2 (\phi_{\gamma} ) (x,y)= 2/ \pi \exp(-2 \, \tilde z)$, and using the decomposition of $\tilde z$ and Proposition \ref{prop:DCfromEfi}.\ref{prop:DCb}.

\revis{Now, let us discuss the computation of the DC decomposition for the inverse multiquadric mother function $\phi(x,y)=(1+x^2+y^2)^{\mu}, \, \, \, \mu<0$. Notice that the decomposition of the terms $\tilde x$ and $\tilde y$ depend only on the PTM and dictionary models (\ref{eq:TransModel}) and (\ref{eq:DictModel}); therefore, they are the same as in the previous case. Since $\phi(\tilde x, \tilde y) = (1+ \tilde z)^\mu$ is a convex function of $\tilde z$ for $\tilde z \geq 0$, the decomposition of $\phi(\tilde x, \tilde y)$ can be obtained using  Proposition \ref{prop:DCfromEfi}.\ref{prop:DCb}. (Although the domain of the function $q$ is $\mathbb{R}$ in Proposition \ref{prop:DCfromEfi}.\ref{prop:DCb}, an examination of the proof in \cite{efiDC} shows that the property can still be applied for a convex function $q$ defined on a domain that is a subset of $\mathbb{R}$.) This gives the decomposition of $U_{\lambda}(\phi_{\gamma})(l)$. The decomposition of $U^2_{\lambda}(\phi_{\gamma})(l)$ can be similarly computed by applying Proposition \ref{prop:DCfromEfi}.\ref{prop:DCb}  to the function $\phi^2(\tilde x, \tilde y) = (1+ \tilde z)^{2\mu}$. Then, the computation of $\tilde E$ is the same as in the previous case.}

%
%

%
%
%
%
%
%

\maketitle{\textit{D. Proof of Proposition \ref{prop:DCclassError}}}\\
\begin{proof}
We rearrange $\tilde E $ in the following form.
\begin{equation}
\begin{split}
\tilde E &= \sum_{m=1}^M   \sum_{i=1}^{N_m} \left( 1+\alpha \, \eta_i^{m }  \right)  \| v^m_i - c^m \, U_{\lambda^m_i} (\phi_{\gamma^m} )  \|^2\\
      & - \sum_{m=1}^M  \sum_{(i,k)\in R^m}  \alpha \, \eta_i^{k} \, \| v_i^{k , m} - c^m \, U_{\lambda^{k , m}_i} (\phi_{\gamma^m} ) \|^2.
\end{split}
\label{eq:appTildeE}
\end{equation}
Equivalently, one can write
\begin{equation*}
\tilde E = \sum_{m=1}^M \tilde {E}^m (\gamma^m, c^m)
\end{equation*}
where
\begin{equation*}
\begin{split}
\tilde {E}^m(\gamma^m, c^m) &=  \sum_{i=1}^{N_m} \left( 1+\alpha \, \eta_i^{m}  \right)  \| v^m_i - c^m \, U_{\lambda^m_i} (\phi_{\gamma^m} ) \|^2\\
& -  \sum_{(i,k)\in R^m}  \alpha \, \eta_i^{k} \, \| v_i^{k , m} - c^m \, U_{\lambda^{k , m}_i} (\phi_{\gamma^m} )  \|^2.
\end{split}
\end{equation*}
From Proposition \ref{prop:DC}.\ref{prop:DCa}, in order to show that $\tilde E$ is a DC function of $\gamma$ and $c$, it is enough to show that $\tilde {E}^m (\gamma^m, c^m)$ is a DC function of $\gamma^m$ and $c^m$ for all $m=1, \, \dots \,, M$.

In the proof of Proposition \ref{prop:DCformTildeE} we have already shown that the squared norm of the difference vector $  \tilde e  = v- c \, U_{\lambda} ( \phi_{\gamma} ) $ is a DC function of $c$ and $\gamma$, and we have described a way to compute its DC decomposition when a DC decomposition of the components of $U_{\lambda} ( \phi_{\gamma})$ is available. Therefore, one can obtain the DC decompositions of the terms $\| v^m_i - c^m \, U_{\lambda^m_i} (\phi_{\gamma^m} ) \|^2$ and $\| v_i^{k , m} - c^m \, U_{\lambda^{k , m}_i} (\phi_{\gamma^m} )  \|^2$ in  the formulation of $\tilde {E}^m (\gamma^m, c^m)$. Then, $\tilde{E}^m (\gamma^m, c^m)$ is a DC function of $\gamma^m$ and $c^m$ as it is a linear combination of these terms, and its decomposition is simply given by Proposition \ref{prop:DC}.\ref{prop:DCa}.

\end{proof}

\end{document}